\documentclass[lettersize,journal]{IEEEtran}
\usepackage{amsmath,amsfonts}
\usepackage{algorithmic}
\usepackage{algorithm}
\usepackage{array}
\usepackage[caption=false,font=normalsize,labelfont=sf,textfont=sf]{subfig}
\usepackage{textcomp}
\usepackage{stfloats}
\usepackage{url}
\usepackage{verbatim}
\usepackage{graphicx}
\usepackage{cite}
\usepackage{newfloat}
\usepackage{listings}
\usepackage{lineno}
\usepackage{float}  
\usepackage{graphicx,subfig}
\usepackage{overpic}
\usepackage{color}
\usepackage{colortbl}
\usepackage[table]{xcolor}
\definecolor{gray9}{gray}{.9}
\floatstyle{ruled}
\floatname{listing}{Listing}
\usepackage{amsmath}
\usepackage{multicol}
\usepackage{multirow}
\usepackage{booktabs}
\usepackage{amsfonts,amssymb} 
\begin{document}

\title{HSIDMamba: Exploring Bidirectional State-Space Models for Hyperspectral Denoising}
% Yang Liu, Jiahua Xiao, Yuan Sun, Yu Guo, Peilin Jiang
\author{Yang Liu, Jiahua Xiao, Xiang Song, Yu Guo, Peilin Jiang, Haiwei Yang, Fei Wang
 \thanks{The authors are with the National Key Laboratory of Human-Machine Hybrid Augmented Intelligence, National Engineering Research Center for Visual Information and Applications, and Institute of Artificial Intelligence and Robotics, Xi’an Jiaotong University, Xi’an, Shaanxi 710049, China, and with the school of software, Xi’an Jiaotong University, Xi’an, Shaanxi 710049, China (e-mail:(1228754216, xjh847286495, songxiang)@stu.xjtu.edu.cn, (yu.guo, pljiang, wfx)@xjtu.edu.cn, yanghw09@163.com) \\
 \textit{(Corresponding author: Yu Guo.)}
 } % <-this % stops a space
% \thanks{This paper was produced by the IEEE Publication Technology Group. They are in Piscataway, NJ.} % <-this % stops a space

}

% \author{Yang Liu, Yantao Ji, Jiahua Xiao, Yu Guo, Peilin Jiang, Haiwei Yang, Fei Wang
% \thanks{The authors are with the National Key Laboratory of Human-Machine Hybrid Augmented Intelligence, National Engineering Research Center for Visual Information and Applications, and Institute of Artificial Intelligence and Robotics, Xi’an Jiaotong University, Xi’an, Shaanxi 710049, China, and with the school of software, Xi’an Jiaotong University, Xi’an, Shaanxi 710049, China(e-mail: (1228754216, jyt1262556482, xjh847286495)@stu.xjtu.edu.cn, (yu.guo, pljiang, wfx)@xjtu.edu.cn, yanghw09@163.com} % <-this % stops a space
% % \thanks{This paper was produced by the IEEE Publication Technology Group. They are in Piscataway, NJ.} % <-this % stops a space

% % \thanks{Manuscript received April 19, 2021; revised August 16, 2021.}
% }

\markboth{Journal of \LaTeX\ Class Files,~Vol.~14, No.~8, August~2021}%
{Shell \MakeLowercase{\textit{et al.}}: A Sample Article Using IEEEtran.cls for IEEE Journals}

\IEEEpubid{This work is licensed under a Creative Commons Attribution 4.0 License. For more information, see https://creativecommons.org/licenses/by/4.0/
}

\maketitle

\begin{abstract}

Effectively modeling global context information in hyperspectral image (HSI) denoising is crucial, but prevailing methods using convolution or transformers still face localized or computational efficiency limitations. Inspired by the emerging Selective State Space Model (Mamba) with nearly linear computational complexity and efficient long-term modeling, we present a novel HSI denoising network named HSIDMamba (HSDM). HSDM is tailored to exploit the capture of potential spatial-spectral dependencies effectively and efficiently for HSI denoising.
In particular, HSDM comprises multiple Hyperspectral Continuous Scan Blocks (HCSB) to strengthen spatial-spectral interactions. HCSB links forward and backward scans and enhances information from eight directions through the State Space Model (SSM), strengthening the context representation learning of HSDM and improving denoising performance more effectively.
In addition, to enhance the utilization of spectral information and mitigate the degradation problem caused by long-range scanning, spectral attention mechanism.
Extensive evaluations against HSI denoising benchmarks validate the superior performance of HSDM, achieving state-of-the-art performance and surpassing the efficiency of the transformer method SERT by $31\%$.
\end{abstract}


\section{Introduction}

\IEEEPARstart{H}{yperspectral} images (HSI) capture highly discriminative surface features over an extensive spectrum and retain complex spectral information enabling its application in diverse tasks, such as remote sensing, object detection, and image analysis~\cite{aburaed2023review,he2023object,jang2024m,yang2023iter,ding2023multi}. Nevertheless, the ubiquity of noise—originating from sensor defects, atmospheric interference, and equipment malfunction, compromises the reliability and fidelity of HSI, impairing applications reliant on accurate spectral data. 

\begin{figure}[t]
\centering
\includegraphics[width=3.4in]{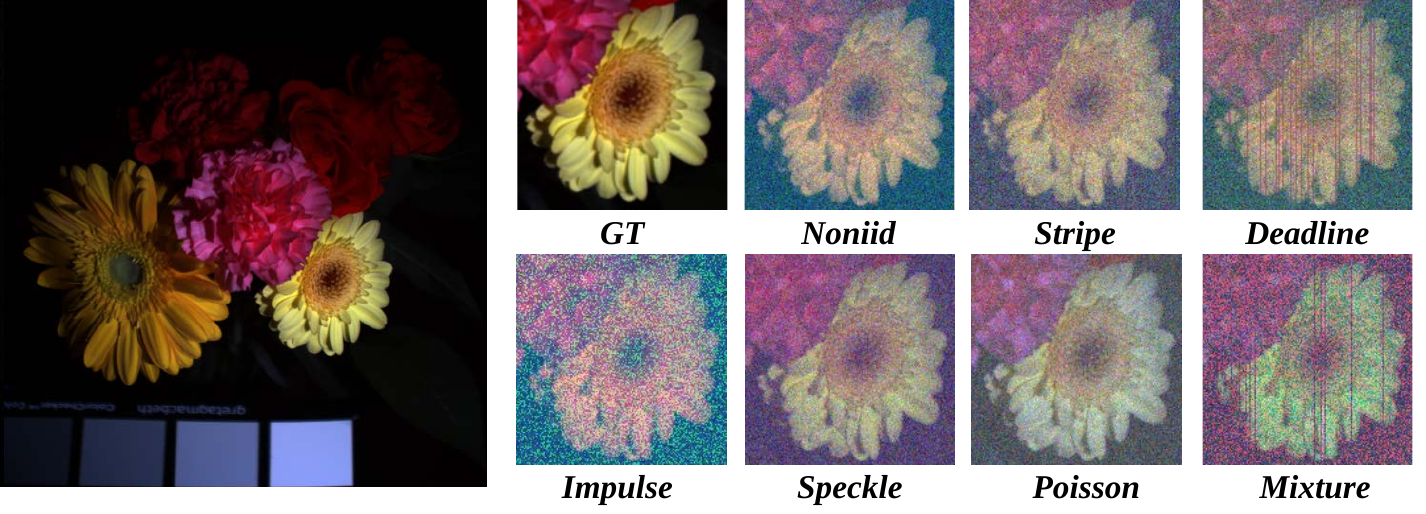}
\caption{Performance of different noises. The visual image is synthesized by bands 29, 19, and 9 of $flowers\_ms$ in CAVE.} 
\label{noise}
\end{figure}

As shown in Figure \ref{noise}, HSI noises manifest in various forms, such as Gaussian, stripe, deadline, impulse, speckle, Poisson noise, and a mixture of the above. As a result of such noise contamination, the integrity of data-driven outcomes, including classification, localization, and analytical tasks, is compromised. This can have detrimental consequences on the veracity and efficacy of decision-making processes relying on HSI. In recognition of these challenges, the imperative to efficiently suppress noise from HSI is underscored.

Hyperspectral images surpass RGB and grayscale imagery with spectral richness but grapple with substantial spectral redundancy and computational demand from contextual dependencies. Despite model-based methods~\cite{sun2018hyperspectral, LLRGTV, NG-Meet} leading to a presumption of hyperspectral noise yield satisfactory results, they require constant fine-tuning of superparameters across different scenes and demand substantial computational and temporal resources, thus constraining their real-world implementation. Convolutional Neural Networks (CNN) based deep learning techniques~\cite{guan2022dnrcnn,qrnn3d} have shown potent potential in HSI denoising, despite inherent challenges such as static adaptability and local restoration biases due to the limited receptive field constrained by kernel size. On the other hand, Transformer architectures effectively addressed the limitations of receptive fields, leading to significant advancements in long-range dependency modeling and model performance~\cite{khan2022transformers,sst,sert}. However, the inherent computational and memory complexity of the self-attention mechanism employed by Transformers results in quadratic growth in computational burden and memory requirements as sequence length increases.

\begin{figure}[t]
\centering
\includegraphics[width=3.4in]{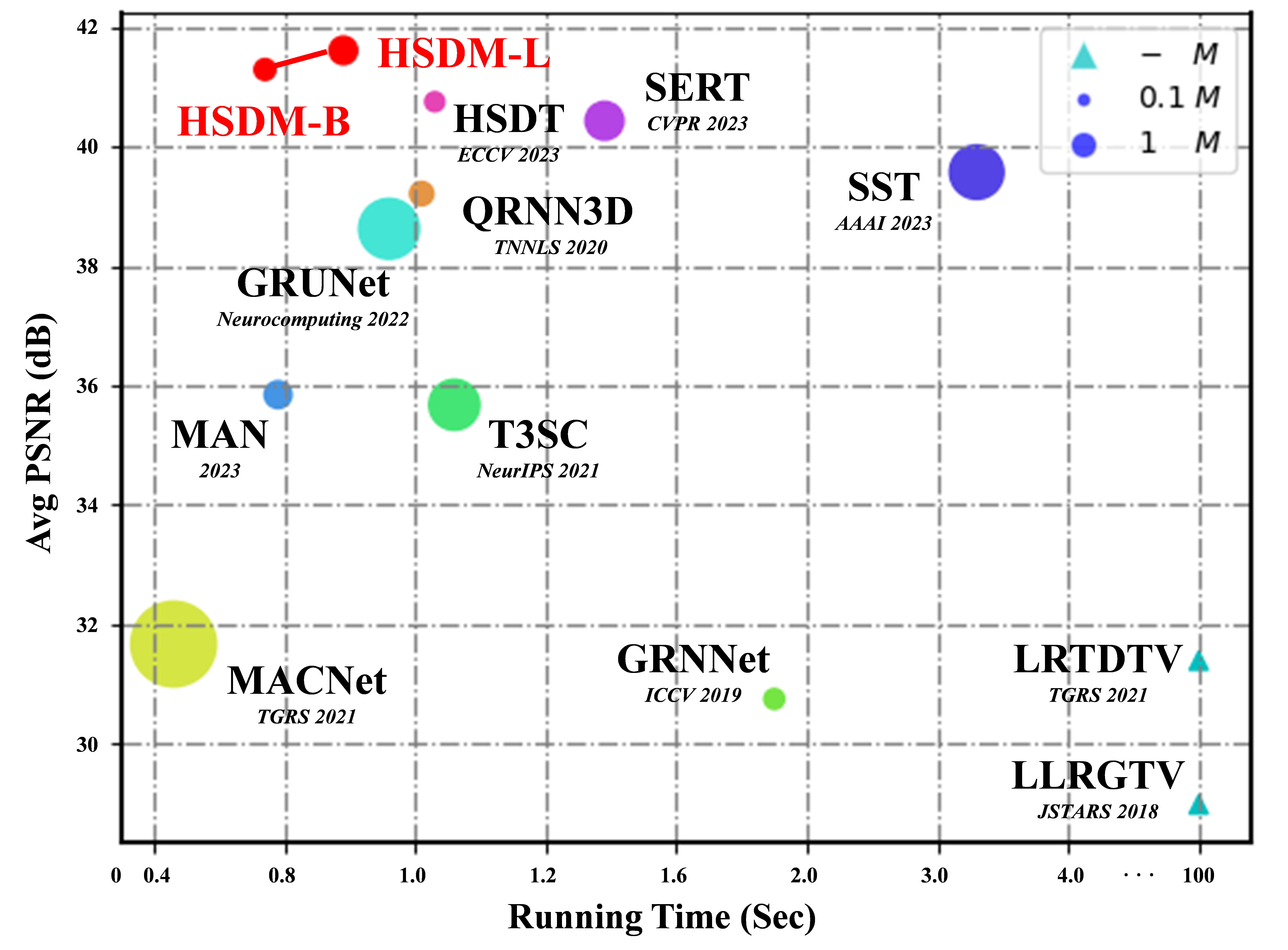}
\caption{Quantitative comparisons (PSNR (dB) vs. Inference Cost(second)) of our HSDM against other deep learning methods on ICVL datasets under mixture noise.} \label{psnr_times}
\end{figure}  

\IEEEpubidadjcol

% While the prospects for HSI restoration based on CNNs are promising, existing work in this domain also encounters several challenges. Firstly, the weight-sharing strategy inherent in CNNs constrains the model's ability to dynamically adapt to different inputs, potentially affecting its performance when dealing with unseen low-quality images. Secondly, the limited receptive field of individual convolutional layers, constrained by the size of the kernel, may significantly increase local restoration biases. Although some studies have proposed different convolution designs, current CNN-based restoration methods are still constrained by the effective receptive field\cite{guan2022dnrcnn,qrnn3d}.

% Subsequent Transformer architectures effectively addressed the limitations of receptive fields, leading to significant advancements in long-range dependency modeling and model performance. However, the inherent computational and memory complexity of the self-attention mechanism employed by Transformers results in quadratic growth in computational burden and memory requirements as sequence length increases\cite{khan2022transformers,sst,sert}.

Recently, significant research interest has been vested in State Space Models (SSMs) due to their potential to confront the challenge of modeling long sequences. They can be viewed as a fusion of recurrent neural networks (RNNs) and CNNs. Notable contributions in this domain include the Structured State Space (S4)~\cite{s4} and its various extensions~\cite{S5, H3, GSS}. The recent development, Mamba~\cite{Mamba}, not only enhances the S4 framework by incorporating a selection mechanism for an input-dependent focus on relevant information but also ensures efficient training and inference through hardware-aware optimizations. Mamba exhibits linear scalability with sequence length, positioning it as a promising alternative for language modeling tasks.
Concurrent research efforts have extended this architecture beyond 1D language tasks to 2D visual domains, including tasks such as image classification and image segmentation~\cite{videomamba, mambair, zigmamba}. Despite this, it is still challenging to directly adapt the original Vision Mamba~\cite{visionmamba} architecture to HSI denoising because the high dimension and abundant spectral data dramatically increase the parameters of the model and demand extensive computational resources. To the best of our knowledge, the potential of leveraging this efficient architecture for addressing HSI denoising challenges has not yet been explored.

% Adapting the original Vision Mamba~\cite{visionmamba} architecture to HSI denoising is challenging due to the high dimensionality and rich spectral data that dramatically increase the model's parameters, demanding extensive computing resources. Modifications such as integrating spectral attention mechanisms can enhance model comprehension of spectral information, consequently reducing computational overhead. Fine-tuning the architecture to encapsulate spectral-spatial dependencies more effectively and incorporating specialized preprocessing steps for unique noise patterns, like striping and deadline noises, could significantly improve the model's noise reduction efficacy and performance.

To address these issues, we propose a network designed for the challenges posed by the inherently long sequences found in HSI. Our solution, termed the Hyperspectral Image Denoising Mamba-based network (HSDM), leverages a selective state space model. HSDM introduces an innovative Hyperspectral Continuous Scanning Block (HCSB) that is meticulously crafted to strengthen spatial-spectral interactions and combat the noise characteristic efficiently. In particular, HSDM effectively overcomes the inherent drawbacks of both convolutional and attention mechanisms. This is achieved by implementing an advanced 2D Selective Scan Module that facilitates a more robust and comprehensive modeling of spatial-spectral long-range dependencies while maintaining linear computational complexity. 

Compared to General Vision Mamba, which typically employs a sweep scanning strategy, HSDM utilizes an alternating continuous scanning mechanism of forward and backward passes. This mechanism achieves comprehensive coverage across all scanning vectors, and in doing so, it enhances computational efficiency and diminishes parameter requirements. The bidirectional continuous scanning mechanism allows for a smooth transition of features before scanning, minimizing potential distortions. Moreover, to further augment the performance of the Mamba network in HSI denoising, we have incorporated the widely proven concept of residual learning. Additionally, we employ a simple yet effective spectral attention module to identify and emphasize pivotal spectral features.
This enhancement has equipped the HSDM to deliver superior noise reduction while preserving the integrity of salient image characteristics, as illustrated in Figure~\ref{psnr_times}.
% HSDM achieves state-of-the-art performance in PSNR and inference speed, demonstrating a well-balanced approach.

% Experimental outcomes across diverse datasets validate the superior performance of our proposed HSDM compared to existing methodologies, evidenced by both objective metrics on synthetic datasets and visual quality assessments on real datasets. 
To summarize, the main contributions of this study are summarized as follows:

\begin{itemize}
    \item We are the first to introduce a simple yet effective Bidirectional State-Space Model for HSI denoising, devising a novel architecture distinct from CNNs and transformers.
    \item To strengthen spatial-spectral interactions and context representation learning, we propose the Bidirectional Continuous Scanning Mechanism to improve detailed information denoising ability effectively and efficiently.
    \item Extensive experiments demonstrate that HSDM achieves state-of-the-art performance and is $31\%$ faster than the transformer method SERT.
    % offering a more balanced solution in terms of performance and efficiency for HSI denoising.
\end{itemize}

\section{Related Work}

In this section, we chiefly explore the latest developments in research on data-driven deep learning techniques. While we refrain from delving into detailed expositions of model-driven approaches, they will be included for comparative analysis in the experimental evaluations. 

\subsection{CNN-based Hyperspectral Denoising}

CNNs effectively extract noise in HSI under various noise conditions, while simultaneously suppressing noise across spatial and spectral dimensions. \cite{zhang2019hybrid} leverage 2D convolutions effectively for HSI denoising, showing strong performance in both spatial and spectral domains. Additionally, 3D convolution significantly enhances the utilization of spatial-spectral structural information. Demonstrated by several techniques, notably in the studies by\cite{yuan2018hyperspectral,guan2022dnrcnn}, 3D convolutions have proven exceptionally adept at capturing the intricacies of spatial-spectral structural information within high-dimensional data. Among the discussed methods, TRQ3DNet\cite{TRQ3DNet} particularly excels in modeling the spatial-spectral dependencies and capturing global spectral correlations, thereby fully leveraging the synergy between spatial and spectral dimensions. Specifically, QRNN3D\cite{qrnn3d} achieves impressive results by combining 3D convolution and recurrent pooling functions to more effectively incorporate spatial-spectral correlations and global contextual information. Nonetheless, despite their formidable noise reduction capabilities, the inherently local receptive fields of CNNs may impede their ability to grasp long-range dependencies in HSI. 

\subsection{ViT-based Hyperspectral Denoising}

Vision Transformer (ViT) has introduced a paradigm shift in the realm of computer vision. Diverging from the traditional confines of CNN architectures, ViT capitalizes on the robust Transformer model, originally conceived for tackling natural language processing challenges. Significantly, the effectiveness of the Transformer's attention mechanism for long-range modeling has been empirically validated in previous works, underscoring its potential utility in a variety of applications. Outfitted with the capacity to process images of varied sizes and demonstrate powerful generalization, ViT surpasses traditional methods in numerous tasks such as image classification, object detection, and segmentation. This paradigm shift mitigates the previously highlighted shortcomings of CNNs.

In recent research into HSI denoising, a variety of innovative methods utilizing the Transformer architecture have been advanced. \cite{sst,sert} introduce enhancements to the foundational attention mechanism of the Transformer model, incorporating design elements specifically tailored to the spectral module. These modifications not only accelerate processing speed relative to the original Transformer but also address spectral distortion and edge blurriness, which are common in long-range modeling of HSI data. \cite{sstd} differentiates itself by mitigating these issues through the preservation of local sensitivity bias, thereby establishing a model that correlates image content for effective HSI denoising. Nevertheless, in spite of determined endeavors to augment performance, the Transformer model encounters constraints in computational and data efficiency due to its dependence on large-scale datasets for effective training, coupled with a computational burden that intensifies when processing high-resolution HSI.

\begin{figure*}[t]
\centering 
\includegraphics[width=7.1in]{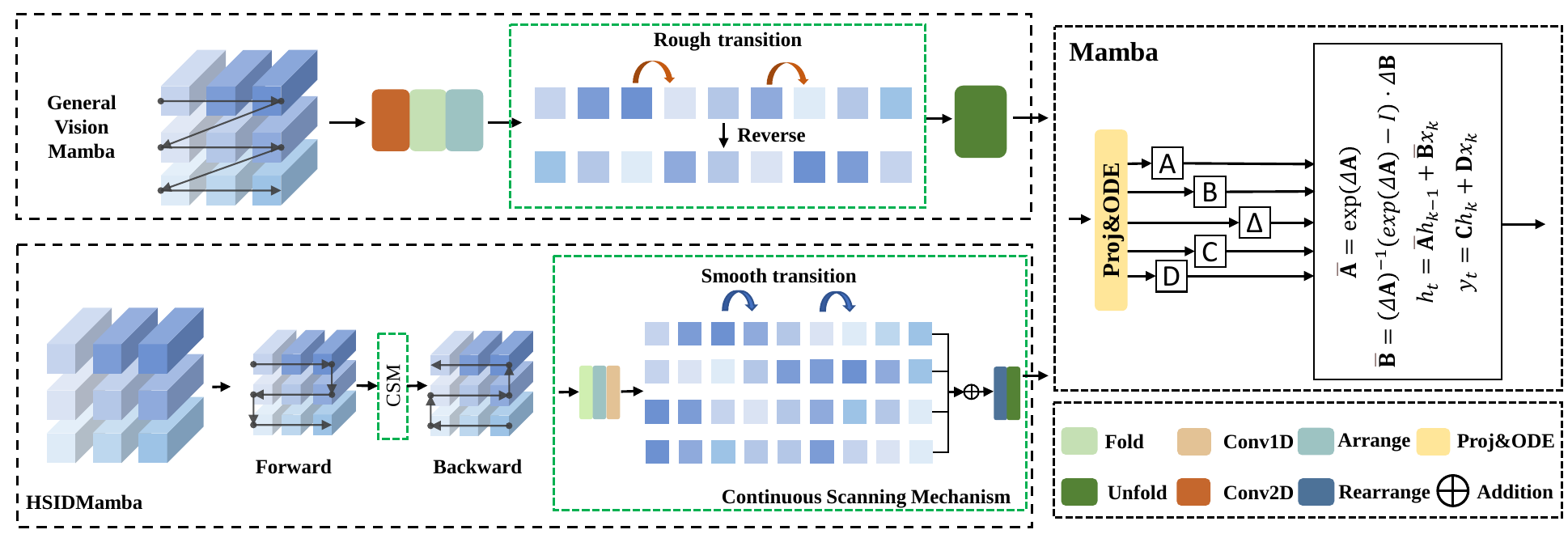}
\caption{Illustration of the 2D-Selective-Scan of General Vision Mamba and HSDM. Compared to generic scanning, our continuous scan strategy achieves smoother transitions by alternatingly merging forward and backward features into the final 2D feature map.}\label{scan}
\end{figure*}   

\subsection{State Space Models}

While CNNs and Transformer models have greatly advanced the field of deep learning through their respective abilities to handle spatial hierarchies and long-range dependencies, researchers continually seek improvements in both architectural efficiency and capability for sequence modeling. It is in this quest for advancement that recent developments in state space models (SSMs) have garnered attention. These models have demonstrated a remarkable proficiency in encapsulating dynamic interconnections and dependencies that are prevalent in language sequences, offering an alternative approach that complements the strengths. S4~\cite{s4}, specifically designed to adeptly manage long-range dependencies while maintaining linear computational complexity, marks a considerable advancement in this area of research. Building upon these improvements, Mamba~\cite{Mamba} introduces an innovative SSM layer capable of dynamically adapting to data. Additionally, it employs a parallel scanning selection mechanism, effectively addressing some of the well-known limitations associated with RNNs and SSMs. Collectively, these developments enable the processing of extensive sequences with linear complexity. This represents a significant advantage over attention-based transformers, particularly in scenarios where managing long-range dependencies is critical.

In the domain of visual processing, SSMs have equally made impactful inroads. Applications of SSMs have been applied to visual tasks such as image classification \cite{hsimamba} as well as in discerning long-range temporal dependencies within movie clip classification scenarios \cite{medmamba}. The advent of methodologies has heralded a period of innovation for state space models. Notably, the Mamba model distinguishes itself with its cutting-edge selective SSM feature, attaining an equipoise of linear computational complexity with the capacity to model long-term dynamics effectively. Our proposed HSDM innovation enhances the original Mamba model by optimizing its data-dependent state space mechanisms for advanced HSI denoising efficiency and effectiveness.

\section{Methods}

In this section, we commence with an introduction to the foundational concepts underpinning state space models. Following this, we delve into an in-depth exploration of the implementation specifics of our proposed HSDM. Finally, we present a succinct analysis and discussion of the computational complexity inherent in various mainstream architectures.

\subsection{Preliminaries}

\textbf{State space models} are statistical models designed to quantify the temporal evolution of a system. In the realm of sequence data, SSMs usually model input data and corresponding hidden states using Ordinary Differential Equations (ODEs). With this unique capability, SSMs have become a pivotal tool in handling temporal data. To further clarify this concept, consider the following example of a linear SSM which can be represented mathematically as follows:
\begin{equation}
    \begin{aligned}
    &{h'}(t) = \textbf{A}h(t) + \textbf{B}x(t) \\
    &y(t) = \textbf{C}h(t) + \textbf{D}h(t)
    \end{aligned} 
    \label{eq1}
\end{equation}
where $x(t) \in {\mathbb{R}^L}$ represents the input sequence, $h(t) \in {\mathbb{R}^N} $ is hidden state, which assist in mapping $x(t)$ to $y(t) \in {\mathbb{R}^M}$. Among the formula, $\textbf{A} \in {\mathbb{R}^{N\times N}}, \textbf{B} \in {\mathbb{R}^{N\times 1}}, \textbf{C} \in {\mathbb{R}^{1\times N}}, \textbf{D} \in {\mathbb{R}}$. N represents the hidden state size.

Following the basic formula (Eq.\ref{eq1}), a discretization process is often employed to integrate these equations. This process can be mathematically represented as:
\begin{equation}
    \begin{aligned}
    &\overline {\textbf{A}} = \exp (\Delta \textbf{A}) \\
    &\overline {\textbf{B}} = {(\Delta \textbf{A})^{ - 1}}(exp(\Delta \textbf{A}) - I) \cdot \Delta {\textbf{B}}
    \end{aligned}
\end{equation}
where $\Delta$ represents the time scale parameter, used to convert continuous parameters $\textbf{A}$ and $\textbf{B}$ into discrete parameters $\overline {\textbf{A}}$ and $\overline {\textbf{B}}$. The zero-order hold (ZOH) method is commonly used for this discretization.

Applying a discretized structure in the neural sequence modeling architectures facilitates the intricate process of learning layer-specific parameters using gradient descent techniques.
Upon applying the discretization process, the continuous equation (Eq.\ref{eq1}), mute to the linearity of the system and finite changes in time, manifests into its discrete counterpart, articulated as:
\begin{equation}
    \begin{aligned}
    &{h_t} = \overline {\textbf{A}} {h_{k - 1}} + \overline {\textbf{B}} {x_k} \\
    &{y_t} = {\mathbf{C}}{h_k} + {\mathbf{D}}{x_k}
    \end{aligned}
\end{equation}

 Mamba has markedly expanded the adaptability of SSM frameworks by attenuating the rigid constraints of time-invariance on the SSM parameters, simultaneously maintaining a high level of computational efficacy. This revolutionary approach has been underscored by numerous studies, illustrating Mamba's proficiency in capturing and modeling long-range dependencies with remarkable accuracy, thereby setting a new benchmark for the effectiveness of SSMs.

\begin{algorithm}[b]
\caption{Learning strategy for HSDM.}\label{algorithm}
\begin{algorithmic}
\STATE 
\STATE \textbf{Input:} $\{ x_1,...,x_n\}\in {{\mathbb R}^{{H} \times {W} \times {B}}}$
\STATE \textbf{Output:} $ \theta $
\STATE \textbf{Hyperparameters:} $Epoch_t, t \in \{0,\propto \}$
\STATE \textbf{ For }$ i = 1,2,...,Epoch_t$
\STATE \hspace{0.5cm}$ B,C,H,W \leftarrow Shape(x_i)$
\STATE \hspace{0.5cm}$ F \leftarrow LayerNorm(Reshape(x,(B,C,L)) $ 
\STATE \hspace{0.5cm} $F_{forward} \leftarrow SSM(rearrange(linear(F)))$
\STATE \hspace{0.5cm}$F_{Backward} \leftarrow SSM(rearrange_b(linear(F)))$
\STATE \hspace{0.5cm}$F_{proj} \leftarrow SiLU(F)$
\STATE \hspace{0.5cm}$ F_{out} \leftarrow F_{proj} \times F_{forward} + F_{proj} \times F_{Backward}  $
\STATE \hspace{0.5cm}$ y_{HMSB} \leftarrow CA(Conv(LN(F+F_{out}))) $
\STATE \hspace{0.5cm}$ y_i \leftarrow x_i + Conv(y_{HMSB})  $
\STATE \hspace{0.5cm}$P(\theta) \leftarrow (y_i,x_i|\theta)$ 
\STATE \hspace{0.5cm}$loss(\theta) \leftarrow - {\textstyle \sum_{j=1}^{n}} logP(\theta )[x_n[j+1],x_n[j]]$
\STATE \hspace{0.5cm}$\theta \leftarrow \theta - \bigtriangledown loss(\theta)$
\STATE \textbf{ return }$\theta$
\end{algorithmic}
\label{alg1}
\end{algorithm}

\subsection{HSID Mamba Model}
\subsubsection{Overall Architecture}
\begin{figure*}[t]
\centering
\includegraphics[width=6.5in]{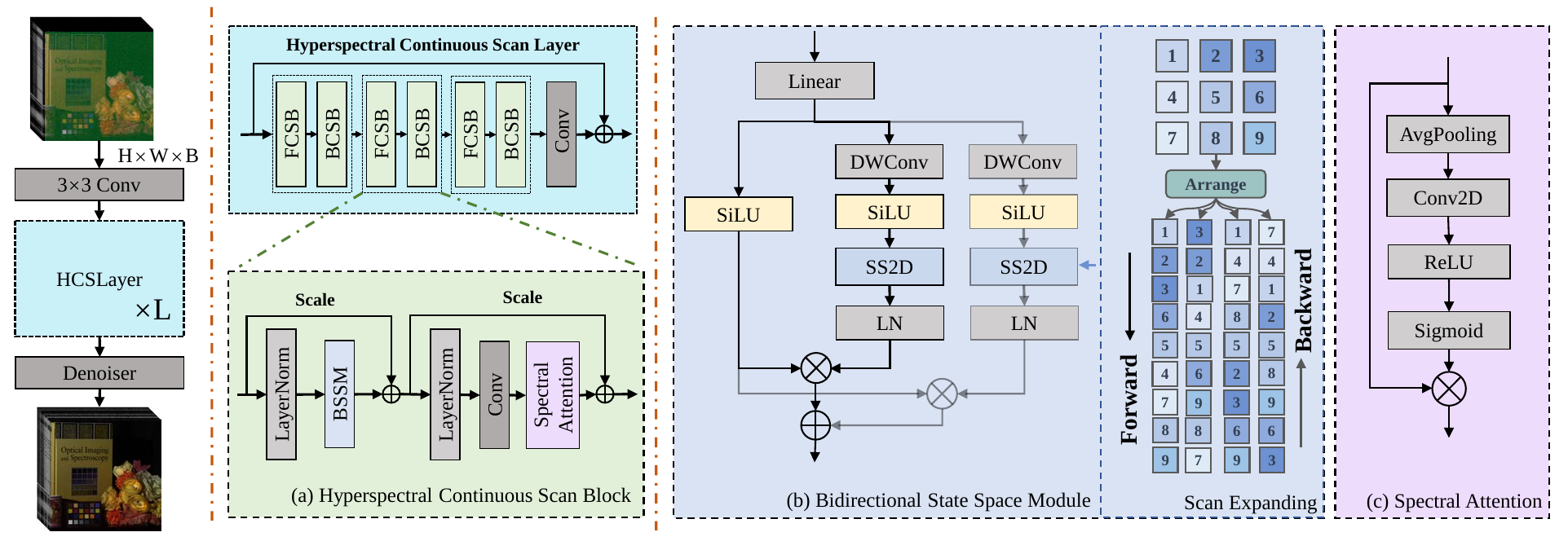}
\caption{Overall framework of our proposed Hyperspectral Denoising Mamba. Adopting the holistic architecture of ViT, we integrate a bidirectional continuous scanning mechanism. (a) Hyperspectral Continuous Scan Block, alternatingly integrating features from forward and backward
scanning directions. (b) BSSM, an augmented State Space module that expands the scanning directions, and (c) spectral attention, to enhance the utilization of spectral information.}\label{network}
\end{figure*}

The architecture of our proposed HSIDMamba (HSDM) model is depicted in Figure \ref{network}, and the overall network algorithm flowchart is shown in Algorithm\ref{alg1}. Starting with an initial shallow feature extractor that teases out preliminary spectral-spatial characteristics from the input HSI. This is followed by an ensemble of Hyperspectral Continuous Scan Layers (HCSL) that form the core of the network. Each HCSL is engineered with a dual-faceted focus: a spatially-aware continuous scanning mechanism to capture the two-dimensional intricacies and a channel attention component that fine-tunes the spectral features for enhanced discriminative learning.
\begin{equation}
    \begin{aligned}
     &F_s = K_1 \otimes Y_in \\
    &F' = HCSL(F_s) \\
    &Y_{out} = Denoiser(F')
    \end{aligned}
\end{equation}
where $Y_in$ represents the input noise HSI, $K_1$ is a 3×3 convolutional kernel, $\otimes$ denotes the convolutional operation, and $F_s,F'$ represents the shallow feature, intermediate feature, respectively. After multi-layer extraction, a Denoiser is used to suppress HSI noise, represented as $Y_{out}$.

Once the HSI input is subjected to the shallow feature extraction phase, the emergent feature maps are channeled into the HCSLs. Within this ambit, a dynamic interaction of scaling and skip-connection strategies facilitates a transformative 1D convolution process. This process is pivotal in conditioning the features for the subsequent Mamba-scan module, which adeptly synthesizes the anisotropic vectors into a coherent representation, conducive to subsequent modeling tasks.

\subsubsection{Bidirectional Continuous Scanning Module}

Incorporating highly efficient parallel scanning, Mamba effectively reduces the limitations of a repetitive sequential nature. However, when addressing hyperspectral data, the original Mamba's unidirectional perception and scanning mechanisms prove inadequate. Recognizing the need to capture the continuity of spatial features, HSDM introduces a Bidirectional Continuous Scanning Module (BCSM). The strategy enhances the scanning process by alternatingly integrating features from the forward and backward scanning directions.

\begin{equation}
    \begin{aligned}
        FS_{i} = \sum_{i=1}^{4} Scan(Direction_{i}, StartPoint_{i})
    \end{aligned}
\end{equation}
where \(FS_{i}\) represents the forward scan from the \(i^{th}\) starting point facilitated by its corresponding scanning direction. For the backward scan, the approach involves inverting the scanning direction relative to each original starting point, thus allowing for a comprehensive bidirectional modeling of the data sequence:

\begin{equation}
    \begin{aligned}
       BS_{i} = FS_{i}(Reverse(Direction_{i}))
    \end{aligned}
\end{equation}

By integrating this bidirectional scanning methodology with the linearized sequence data, the approach not only significantly enhances the inductive bias for simulating the similarities across multidimensional data but also notably curtails both inference duration and memory demands. The entire scanning method is illustrated in Figure \ref{scan}, providing a visual representation of the process.

\subsubsection{Hyperspectral Continuous Scan Block}

HCSL consists of multiple alternately Hyperspectral Continuous Scan Blocks (HCSBs), each with dual scaling to selectively retain feature information. Inside the HCSL, the preserved features still hold their 2D form. Data undergo normalization and reordering layers for refinement, transforming features into a one-dimensional format. Before entering the Bi-directional State Space Model (BSSM) layer and the spectral attention mechanism, these one-dimensional data go under layer normalization for consistent scaling and better spectral dependency capture. The process can be expressed as:
\begin{equation}
    \begin{aligned}
     &F_{tmp} = BSSM(LN(F_{in})) + Scale(F_{in}) \\
    &F_{out} = CA(Conv(LN(F_tmp))) + Scale(F_{tmp})
    \end{aligned}
\end{equation}
where scale operation here is realized by 3*3 convolution.

As depicted in Figure \ref{scan}, we aggregated the results obtained from various scans, employing two rounds of normalization. For an intermediate feature derived from HCSB, denoted as $F_i'$, the subsequent output feature $F_{i+1}'$ of the HCSB can be precisely articulated as follows:
\begin{equation}
    \begin{aligned}
     &F_j = SiLU(DWConv(Fold(rearrange(F_i')))) \\
    &{F_{{\text{i}} + 1}} = UnFold(LN(\sum\limits_{j = 0}^t {LN(SS2D({F_{\text{j}}})} )))
    \end{aligned}
\end{equation}
where $F_j$ represents the reordered 2D features, and t represents the number of rearranged sequences. When $i$ is odd, it indicates forward scanning, when $i$ is even, it signifies backward.

\subsection{Complexity Analysis}

In previous work, CNNs and Transformers played important roles in processing sequence and image data. However, their time complexities present optimization challenges. For CNNs, complexity grows with input image size and filter count—optimizations can include using separable convolutions to reduce parameters. Transformers exhibit quadratic complexity relative to input length but can be optimized through attention mechanisms like sparse attention. Our proposed HSDM, on the contrary, benefits from linear complexity, mitigating scalability issues inherent to the other architectures. This subsection aims to analyze the time complexity across these base architectures.

\textbf{Convolutional neural networks}: 
The computational complexity of the conventional Convolutional Neural Networks (CNNs) architecture can be succinctly expressed as follows:
\begin{equation}
    \Omega (T)=\Omega(K^2\times H\times W\times C)
\end{equation}
where k is convolutional kernel size, \(H,W,C\) denotes the input feature dimension.

\textbf{Transformer with Self-Attention}: 
The computational complexity of the conventional global self-attention transformer architecture can be mathematically represented as follows:
\begin{equation}
    \Omega (T)=\Omega(H\times W\times C^2)
\end{equation}

\textbf{Transformer with Cross-Attention}: 
The formula describing the computational complexity of the Cross-Attention transformer architecture can be succinctly expressed as follows:
\begin{equation}
    \Omega (CSSA)=(\frac{H^2W^2\times C^2}{h\times w})
\end{equation}
where \(W\) denotes the input feature windows with sizes {[h, w]}.

\textbf{HSID Mamba}: The formula for HSDM architecture can be expressed as:
\begin{equation}
    \Omega (HSIDM)=( T\times H\times W\times C)
\end{equation}
where ${\rm{T}}( \ll C)$ represents the number of times to rearrange 2D features.

Therefore, when dealing with lengthy sequence data, HSDM offers distinct advantages over conventional CNNs and Transformers. HSDM excels in handling extended sequence data with greater efficiency, all while minimizing the computational overhead typically associated with HSI Denoising.

\section{Experiments}

\begin{table*}[!t]
\centering
\caption{Quantitative comparison on ICVL datasets with Gaussian noise levels. Best in \textbf{bold}, second best \underline{underlined}.} \label{icvl}
        \renewcommand\arraystretch{1.3}
        \tabcolsep=3pt
        % \resizebox{\linewidth}{!}{
		% \newcolumntype{C}{>{\centering\arraybackslash}X}

		\begin{tabular}{l|ccc|ccc|ccc|ccc}
                \hline
                \multirow{2}{*}{Methods} & \multicolumn{3}{c|}{10} & \multicolumn{3}{c|}{30} & \multicolumn{3}{c|}{50} & \multicolumn{3}{c}{70} \\
                & PSNR & SSIM & SAM & PSNR & SSIM & SAM & PSNR & SSIM & SAM & PSNR & SSIM & SAM \\
                \hline 
                Noisy &28.13 	&0.8792 	&0.3267 	&18.59 	&0.5523 	&0.6622 		&14.15 	&0.3476 &0.8554 		&11.23 	&0.2301 	&0.9851 	\\ 
                LRTDTV~\cite{LRTDTV} & 40.05  & 0.9642  &0.1523  &37.58 & 0.9414  &0.1137  &35.68  & 0.9180 & 0.1756  & 34.22  & 0.8951  & 0.2534   \\
                LLRGTV~\cite{LLRGTV} &42.30  & 0.9864  & 0.1256  & 36.36  & 0.9494  & 0.1394  & 33.06  & 0.9015  & 0.2756  & 30.76  & 0.8495  &  0.3283  \\
                NGMeet~\cite{NG-Meet}&42.42&0.9903&0.1207&37.64&0.9495&0.1104&35.92&0.9387&0.1698&35.21&0.9046&0.2378 \\
                \midrule
                QRNN3D~\cite{qrnn3d} & 44.75  & 0.9972  & 0.0370  & 42.50   & 0.9959  & 0.0359  & 40.45  & 0.9936  & 0.0409  & 38.33  & 0.9902  &0.0500  \\
                GRNet~\cite{GRN} & 36.12  & 0.9764  & 0.0933  & 36.17 & 0.9796 & 0.0888  & 29.64  & 0.9220  & 0.2174  &29.64  & 0.9220   & 0.2174  \\
                MACNet~\cite{xiong2021mac} & 41.30   & 0.9945  & 0.0540   & 38.45  & 0.9903  & 0.0818  & 35.36  & 0.9846 & 0.1000  & 32.70 & 0.9758  & 0.1205  \\
                T3SC~\cite{bodrito2021trainable} & 43.70  &0.9961  &0.0449  & 40.62  & 0.9934  & 0.0514 & 37.87  & 0.9878  & 0.0675  & 35.17  & 0.9776  & 0.1042  \\

                GRUNet~\cite{GRUNet}&45.06&0.9972&0.0339&41.67&0.9953&0.0390&39.50&0.9924&0.0464&36.17&0.9865&0.0663 \\
            MAN~\cite{man} &40.48&0.9823&0.0705&37.29&0.9762&0.0989&35.16&0.9678&0.1289&33.49&0.9573&0.1584 \\

                SST~\cite{li2023spatial} & 44.75  & 0.9972  & 0.0370  & 42.50   & 0.9959  & 0.0359  & 40.45  & 0.9936  & 0.0409  & 38.33  & 0.9902  &0.0500  \\
                SERT~\cite{li2023spectral} & 46.03  & 0.9981  & 0.0342 & 42.92 & 0.9964  &\textbf{0.0334}  & 40.64  & 0.9940  & \textbf{0.0390}  & 38.69  & 0.9910  & 0.0477 \\
                HSDT~\cite{hsdt} & \underline{47.04}  & \underline{0.9985}  & \underline{0.0260} & \underline{42.95}   & 0.9964  & 0.0341  & 40.76  &  0.9942  & 0.0400  & 39.24  & 0.9918  &0.0455  \\
                \midrule
                 HSDM-B  & 46.31 & 0.9982 & 0.0289  & 
                 42.74  & \underline{0.9964}  &0.0362 & 
                 \underline{40.77} &\underline{0.9943}  &0.0408 & 
                 \underline{39.39} &\underline{0.9922} &\underline{0.0447} \\
                 HSDM-L  & \textbf{47.28} & \textbf{0.9984} & \textbf{0.0248}  & 
                 \textbf{43.15}  & \textbf{0.9965}  &\underline{0.0335} & 
                 \textbf{40.99} &\textbf{0.9946}  &\underline{0.0399} & 
                 \textbf{39.65} &\textbf{0.9924} &\textbf{0.0423} \\
                \hline
            
            \end{tabular}
        % }
\end{table*}

In this section, we explore synthetic datasets that are corrupted by mixed noise, real datasets, and remote datasets. We extensively assess our proposed HSDM and conduct experiments to analyze its efficacy thoroughly.

\subsection{Overall Network Architecture}

As illustrated in Figure \ref{network}, in the baseline configuration of HSDT, we employed 3 layers of HCSL, each comprising 4 Blocks for alternating forward and backward Continuous Scan. Additionally, before the modules, a residual connection and feature concatenation from the module output were utilized. Finally, skip connections were employed to add the initial input of each layer to the overall layer output, ensuring the stability of the denoising pattern.

\subsubsection{Training Datasets}

In training our proposed HSDM, we adopt a dataset configuration of \cite{wei20203}, where 100 images are randomly sampled from the ICVL dataset \cite{arad2016sparse}. These images were centrally cropped to achieve a spatial resolution of 512×512 with 31 spectral bands. Further adjustments are made to the spatial resolution of the images to 64×64 through operations including random cropping, rotation, and flipping, resulting in approximately 53,000 training samples.

\subsubsection{Testing Datasets}

Our experimental setup encompasses datasets including ICVL and CAVE \cite{CAVE}, representing synthetic scenarios. Specifically, for the ICVL test set, we randomly selected 50 unique images that were not part of the training set to assess model performance. In the case of the CAVE dataset, 30 images were meticulously chosen for testing purposes, each boasting a spatial resolution of 512×512 pixels and comprising 31 spectral bands, ensuring a comprehensive evaluation across diverse spectral characteristics and spatial attributes.

To evaluate the model's ability to suppress noise, testing datasets also encompass remote sensing datasets including the Washington DC dataset, real datasets\cite{sert} and Urban \cite{urban}.

% \footnote{\url{https://engineering.purdue.edu/~biehl/MultiSpec/hyperspectral.html}}

\subsubsection{Comparison Methods and Evaluation Metrics}

Our experiments encompass a comparative analysis between HSDM and a spectrum of methodologies, including traditional model-based approaches including LRTDTV \cite{LRTDTV} and LLRGTV \cite{LLRGTV}, as well as deep learning-based methods such as GRNet \cite{kuang2019fashion}, MACNET \cite{xiong2021mac}, T3SC \cite{bodrito2021trainable}, GRUNet\cite{GRUNet}, MAN~\cite{man}, SST \cite{li2023spatial}, SERT \cite{li2023spectral}, and HSDT \cite{hsdt}. In particular, SST, SERT, and HSDT are Transformer-based methods, and others are hybrid-structured convolutional neural networks. To comprehensively evaluate the performance of these methodologies, we employ three distinct image quality evaluation metrics: Peak Signal-to-Noise Ratio (PSNR), Structural Similarity (SSIM) \cite{SSIM}, and Spectral Angle Mapper (SAM) \cite{SAM}. Through this meticulous comparison, we aim to elucidate the superior denoising capabilities of HSDM in various real-world scenarios.

\begin{table*}[t]
\centering
\caption{Quantitative comparison on various datasets under mixture noise. Best in \textbf{bold}, second best \underline{underlined}.} \label{icvl_com}
            \renewcommand\arraystretch{1.3}
            \tabcolsep=3pt
		% \resizebox{\linewidth}{!}{
		\begin{tabular}{l|ccc|ccc|ccc|ccc|ccc}
            \hline
            \multirow{2}{*}{Methods} & \multicolumn{3}{c|}{Noniid} & \multicolumn{3}{c|}{Stripe} & \multicolumn{3}{c|}{Deadline} & \multicolumn{3}{c|}{Impulse} & \multicolumn{3}{c}{Mixture}\\
            & PSNR & SSIM & SAM & PSNR & SSIM & SAM & PSNR & SSIM & SAM & PSNR & SSIM & SAM& PSNR & SSIM & SAM\\
            \hline 
            Noisy &18.04 	&0.5056 &0.8116	&17.42	&0.4826 &0.8258 &17.60 &0.4852 	&0.8334	&14.99 	&0.3811	&0.8795	&13.97&0.3392&0.8987\\ 
            % LLRT~\cite{LLRT}   & 24.79  & 0.6294  &  0.4889  &26.60  & 0.8832  & 0.1004  & 16.63  & 0.4895 & 0.8437 & 22.80  & 0.5083 & 0.5718 \\
            LRTDTV~\cite{LRTDTV}  & 37.62  & 0.9434   & 0.1113   & 36.74   & 0.9375  & 0.1207  & 35.67   & 0.9289   & 0.1327  & 36.73  & 0.9373  & 0.1307  &34.46 &0.9184 &0.1431   \\
            % NMOG~\cite{NG-Meet} & 34.85 & 0.9427 & 0.0412 &34.25 & 0.9309 & 0.0508 & 33.48 & 0.9282 & 0.0437 & 30.00 & 0.8407 & 0.0348 \\
            LLRGTV~\cite{LLRGTV}  & 35.29  & 0.9381 &  0.1087   &34.99  & 0.9352  & 0.1130  & 33.67  & 0.9186  & 0.1234  &33.12  & 0.8958  & 0.1247 &31.39 &0.8756 &0.1406 \\
            NGMeet~\cite{NG-Meet}  & 34.85   & 0.9427   &  0.1412   &34.25   &0.9309  & 0.1508  & 33.48   & 0.9282  & 0.1437  & 30.00  & 0.8407  &0.1448 &28.98 &0.8333 &0.1431  \\
            \midrule
            QRNN3D~\cite{qrnn3d} &42.06  & 0.9948  & 0.0501  &41.63  & 0.9943   & 0.0525  &41.74   & 0.9947   & 0.0514   &40.35   &0.9915   & 0.0741 &39.22 & 0.9904& 0.0809 \\
            GRNet~\cite{GRN} & 34.55  &0.9757  & 0.0937  & 34.18  & 0.9743  & 0.0989 & 33.13 & 0.9715  & 0.0969  &32.46  & 0.9575  & 0.1461 &31.67 &0.9557 & 0.1431  \\
            MAC-Net~\cite{MACNet} & 39.84  & 0.9922  & 0.0764 & 39.01  &0.9909  &0.0730 & 36.95  & 0.9873  & 0.0966  & 34.56 & 0.9555  & 0.1899 &30.75 &0.9332 &0.2673   \\
            T3SC~\cite{T3SC} &41.83  & 0.9944 & 0.0525  & 41.24  & 0.9937  & 0.0583 & 39.54  & 0.9923  &0.0856  & 37.94 & 0.9836 & 0.1151 &35.68 &0.9790 &0.1389  \\
            GRUNet~\cite{GRUNet}&42.77&0.9962&0.0401&42.12&0.9956&0.0430&42.09&0.9956&0.0424&41.18&0.9943&0.0528&38.63&0.9913&0.0651 \\
            MAN~\cite{man} &38.84&0.9769&0.0919&38.39&0.9762&0.0964&38.27&0.9773&0.0938&36.75&0.9703&0.1376&35.85&0.9691&0.1457 \\
            SST~\cite{sst} & 43.59 & 0.9966 & 0.0354  & 43.11  & 0.9963 & 0.0366 &42.98  & 0.9960 & 0.0372  & 42.18  &0.9952 &0.0424 &39.58 &0.9928 &0.0480  \\
            SERT~\cite{sert} &\underline{43.65}  & \underline{0.9969} & \textbf{0.0343} &\underline{43.18}  & \underline{0.9967} &\textbf{0.0359}& \underline{43.24}  & \underline{0.9967}  &\underline{0.0362} & 42.25 & 0.9953 & 0.0420  &40.44 &0.9940 &0.0470  \\
            HSDT~\cite{hsdt} & 43.40 & 0.9965 & 0.0372 &42.96  & 0.9962&0.0400 & 43.02 & 0.9963 &0.0391& 42.27 & 0.9956 & 0.0442 &40.76&0.9940& 0.0505 \\
            \midrule
             HSDM-B &43.44 & 0.9967 & 0.0371 & 43.03 & 0.9964 & 0.0388 & 42.99 & 0.9964 & 0.0391 & \underline{42.43} & \underline{0.9958} & \underline{0.0418}  &\underline{41.30}&\underline{0.9950}&\underline{0.0459}\\
             HSDM-L &\textbf{43.84} & \textbf{0.9972} & \underline{0.0347} & \textbf{43.25} & \textbf{0.9969} & \underline{0.0366} & \textbf{43.23} & \textbf{0.9970} & \textbf{0.0354} & \textbf{42.55} & \textbf{0.9961} & \textbf{0.0396}  &\textbf{41.62}&\textbf{0.9954}&\textbf{0.0431}\\
            \hline
            \end{tabular}
            % }
\end{table*}

\def\fwidth{0.19\linewidth}
\def\arraystretch{0.5}
\tabcolsep=1.5pt
\begin{figure*}[t]
\centering
\begin{tabular}{ccccc}
     \includegraphics[width=\fwidth]{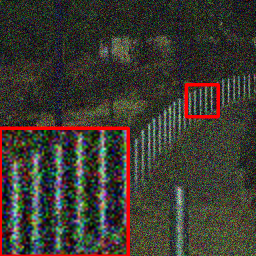}&
     \includegraphics[width=\fwidth]{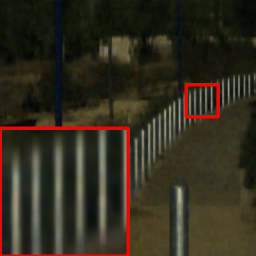} &  
       \includegraphics[width=\fwidth]{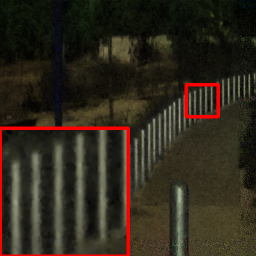}&
    \includegraphics[width=\fwidth]{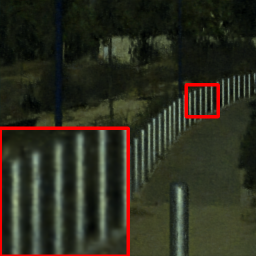}&
     \includegraphics[width=\fwidth]{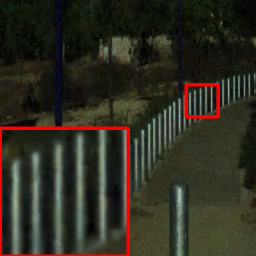}  
     \\ 
      \scriptsize{Noisy}
     & \scriptsize{LRTDTV}
      & \scriptsize{LLRGTV}
    & \scriptsize{MAC-Net}
      &  \scriptsize{T3SC}
     \\
      \includegraphics[width=\fwidth]{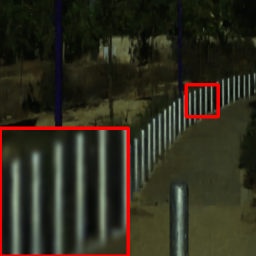}&
     \includegraphics[width=\fwidth]{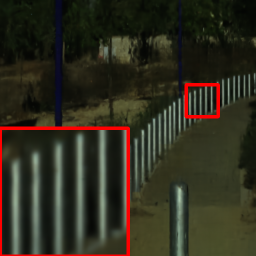} & 
     \includegraphics[width=\fwidth]{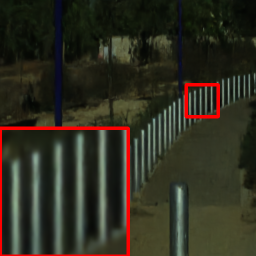} &  
     \includegraphics[width=\fwidth]{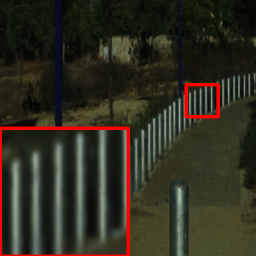}&
     \includegraphics[width=\fwidth]{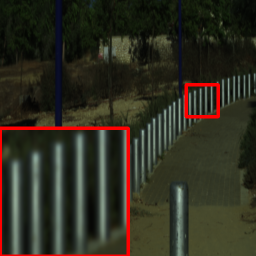} 
    \\
      \scriptsize{SST}
     & \scriptsize{SERT}
     &\scriptsize{HSDT}
      & \scriptsize{\textbf{HSDM-B}}
     & \scriptsize{GT}
\end{tabular}
\caption{Visual comparison of denoising on the ICVL dataset under Gaussian noise levels. The HSI bands 29, 19, and 9 combine to create the visual image.}
\label{fig: icvl}
\end{figure*}

\subsubsection{Noise Settings}

We conduct experiments by adding various levels of Gaussian noise and complex noise combinations to clean images. Each type of noise is applied with specific settings as outlined below:

\begin{itemize}
\item Non-i.i.d. Gaussian Noise: Zero-mean Gaussian noise, with intensities ranging randomly from 10 to 70, is added to each spectral band of the HSI.
\item Non-i.i.d. Gaussian + Stripe Noise. In addition to the Gaussian noise, stripes are introduced randomly to one-third of the bands in the column. The proportion of stripes within each selected band varies from $5\%$ to $15\%$.
\item Non-i.i.d. Gaussian + Deadline Noise. Similar to Gaussian + Stripe Noise, but with deadline noise replacing stripe noise.
\item Non-i.i.d. Gaussian + Impulse Noise. Gaussian noise is added to each band as in Non-i.i.d. Gaussian Noise, and impulse noise with intensities ranging from $10\%$ to $70\%$ is introduced randomly to one-third of the bands.
\item Mixture Noise. Combines all the noise settings from the previous setting, randomly applied to each band.
\end{itemize}

\subsubsection{Implementation Details}

To align with the training configurations of mainstream approaches, for the training on Gaussian noise, we employed a mixture of Gaussian noises at varying levels. For complex noise, we introduced a random overlay of the first four types of noises, thereby enhancing the network's adaptability and recovery capabilities. Additionally, HSDM utilizes the Adam\cite{adam} optimizer to minimize the mean squared error, following a training strategy similar to that outlined in \cite{wei20203}, with minor modifications in the learning rate schedule. Specifically, the learning rate undergoes three stages: initially set to 3e-4, it is then reduced by a factor of 5 after every 20 epochs. Training is performed with a batch size of 8 on an RTX3090 for 100 epochs.

\subsection{Efficiency analysis}

In this subsection, we analyze the performance of different methods regarding model parameter size and time consumption cost on the CAVE dataset with noise case 5. All test experiments are conducted on an RTX3090 GPU, and the results are presented in Table \ref{effcient}.

\begin{table}[t]
    \caption{Complexity comparisons on the CAVE dataset with dimensions of 512×512×31. Best in \textbf{Blod}, second best \underline{Underlined}.}
    \renewcommand\arraystretch{1.3}
    \tabcolsep=3pt
    \centering
        \begin{tabular}{l|cc|ccc}
        \hline 
        Methods & Params(M) & Times(s) & PSNR&SSIM& SAM\\
        \hline
        \multicolumn{1}{l|}{LLRT~\cite{LLRT}} & \multicolumn{1}{c}{-}  &807.21 &24.59  & 0.5795   &0.8171  \\	
        \multicolumn{1}{l|}{LRTDTV~\cite{LRTDTV}} & \multicolumn{1}{c}{-}&97.65 &33.82 &	0.9085 	&\textbf{0.2938}   \\
        \multicolumn{1}{l|}{LLRGTV~\cite{LLRGTV}} & \multicolumn{1}{c}{-} & 171.72 & 27.12 &	0.7221 &0.6567   \\
        \multicolumn{1}{l|}{NGMeet~\cite{NG-Meet} }& \multicolumn{1}{c}{-}  & 86.03 & 23.20   & 0.6229   & 0.7874   \\
        \midrule
        \multicolumn{1}{l|}{QRNN3D~\cite{qrnn3d}} & 0.86 &1.01 & \textbf{36.55}  & 0.9825    & 0.4244   \\
        \multicolumn{1}{l|}{GRNet~\cite{GRN}} & 41.44 &\textbf{0.63} &28.44  &0.8899 & 0.6329  \\
        \multicolumn{1}{l|}{MACNet~\cite{MACNet}} & \textbf{0.43}  &3.09  & 28.53   & 0.8920   & 0.6234   \\
        \multicolumn{1}{l|}{T3SC~\cite{T3SC}} & 0.83  &1.06 & 33.61  & 0.9728   &0.4137  \\
        \multicolumn{1}{l|}{GRUNet~\cite{GRUNet}} &14.28&0.96&35.98  & 0.9764  & 0.4049  \\
        \multicolumn{1}{l|}{MAN~\cite{man}} &0.87&0.79&35.15  & 0.9712 & 0.4990  \\
        \midrule
        \multicolumn{1}{l|}{SST~\cite{sst}} & 4.10 & 3.36 & 34.89 & 0.9616  & 0.4095  \\
        \multicolumn{1}{l|}{SERT~\cite{sert}} & 1.91 &1.29& 35.86   & 0.9737   &0.3403 \\
        % \multicolumn{1}{l|}{HSDT~\cite{hsdt}} &\textbf{0.13}  &1.02& 38.52   & 0.9892 &0.2362 \\
        \midrule
        \multicolumn{1}{l|}{HSDM-B}  & \underline{0.68} & \underline{0.77}&35.99 &\underline{0.9867} & 0.3221 \\
        \multicolumn{1}{l|}{HSDM-L}  & 1.08 & 0.89&\underline{36.14} &\textbf{0.9869} & \underline{0.3014} \\
        \hline
        \end{tabular}
    \label{effcient}
\end{table}

HSDM showcases outstanding performance through its implementation of the Transformer-based architecture, striking a remarkable balance between computational efficiency and parameter optimization. Unlike other deep learning methodologies, it boasts accelerated computation times while maintaining a commendable equilibrium between resource utilization and performance metrics.

\def\fwidth{0.095\linewidth}
\def\arraystretch{0.5}
\renewcommand{\tabcolsep}{0.5 pt}
\begin{figure*}[t]
\centering
\begin{tabular}{cccccccccc}  
     \includegraphics[width=\fwidth]{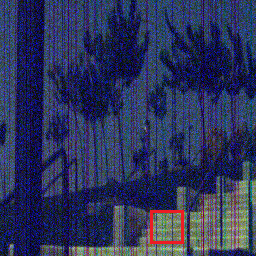} 
    & \includegraphics[width=\fwidth]{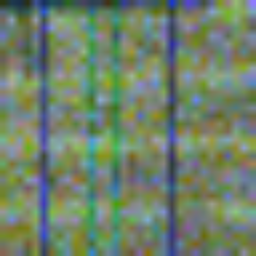}&
     \includegraphics[width=\fwidth]{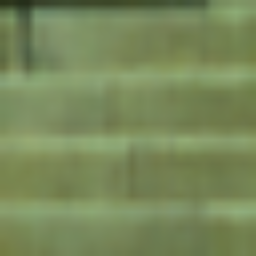} &  
       \includegraphics[width=\fwidth]{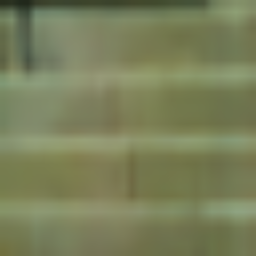}&
    \includegraphics[width=\fwidth]{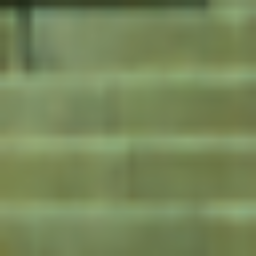}&
     \includegraphics[width=\fwidth]{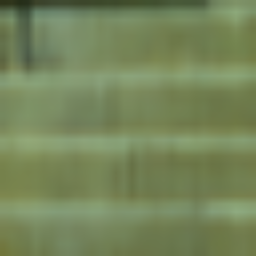}  
     &\includegraphics[width=\fwidth]{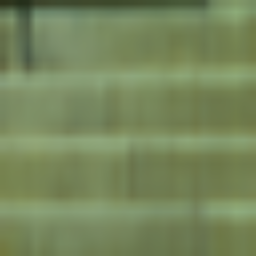}&
     \includegraphics[width=\fwidth]{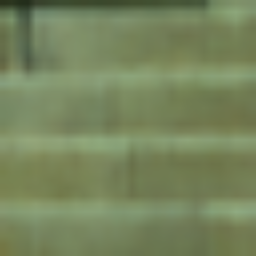} & 
     \includegraphics[width=\fwidth]{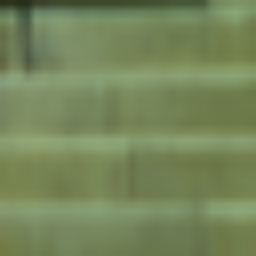}&
      \includegraphics[width=\fwidth]{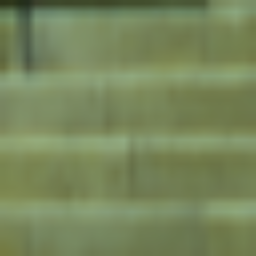} 
     \\ 
     \includegraphics[width=\fwidth]{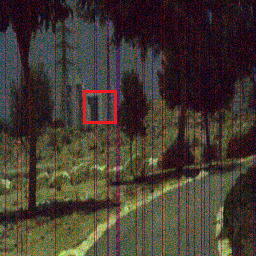} 
    & \includegraphics[width=\fwidth]{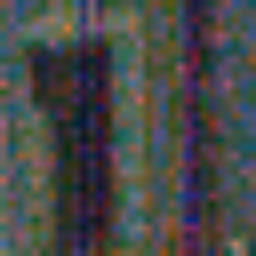}&
     \includegraphics[width=\fwidth]{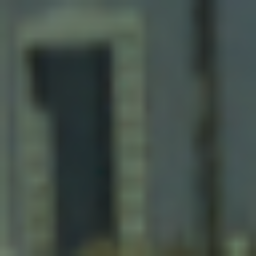} &  
       \includegraphics[width=\fwidth]{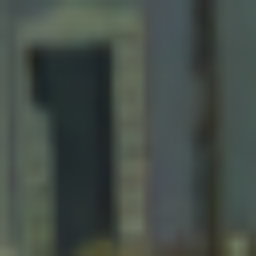}&
    \includegraphics[width=\fwidth]{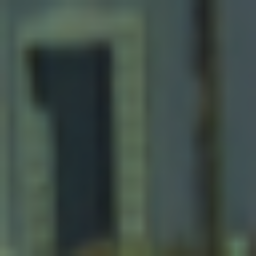}&
     \includegraphics[width=\fwidth]{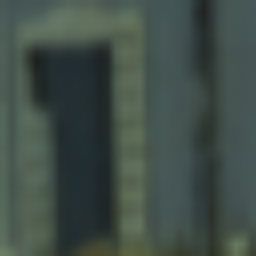}  
     &\includegraphics[width=\fwidth]{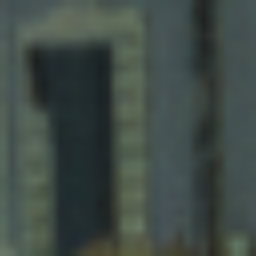}&
     \includegraphics[width=\fwidth]{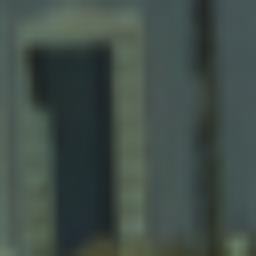} & 
     \includegraphics[width=\fwidth]{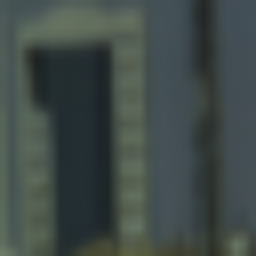}&
      \includegraphics[width=\fwidth]{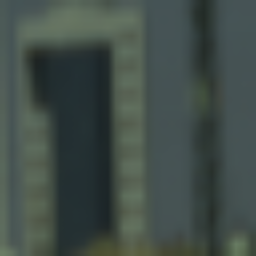} 
     \\ 
     \includegraphics[width=\fwidth]{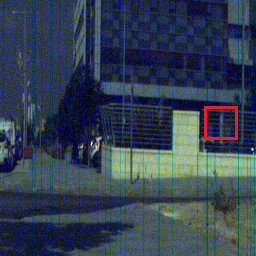} 
    & \includegraphics[width=\fwidth]{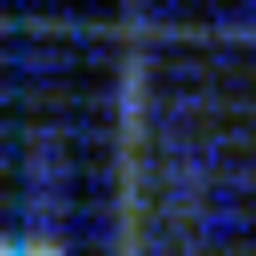}&
     \includegraphics[width=\fwidth]{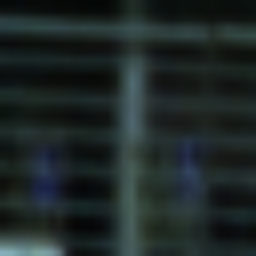} &  
       \includegraphics[width=\fwidth]{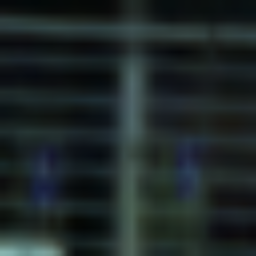}&
    \includegraphics[width=\fwidth]{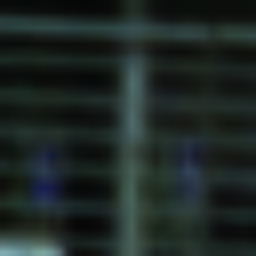}&
     \includegraphics[width=\fwidth]{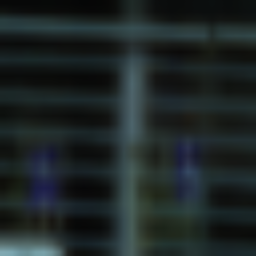}  
     &\includegraphics[width=\fwidth]{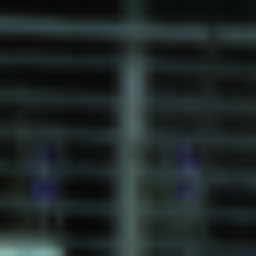}&
     \includegraphics[width=\fwidth]{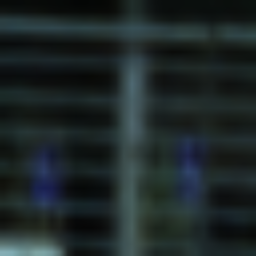} & 
     \includegraphics[width=\fwidth]{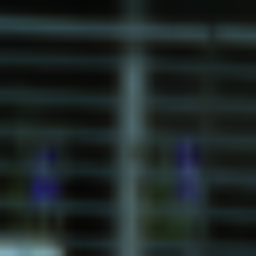}&
      \includegraphics[width=\fwidth]{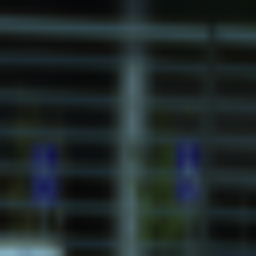} 
     \\ 
     \includegraphics[width=\fwidth]{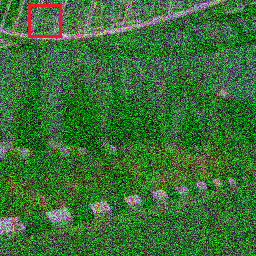} 
    & \includegraphics[width=\fwidth]{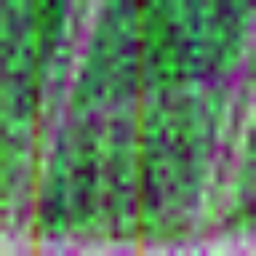}&
     \includegraphics[width=\fwidth]{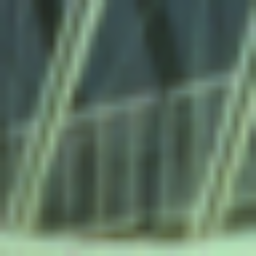} &  
       \includegraphics[width=\fwidth]{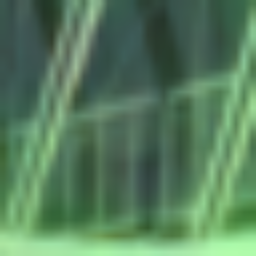}&
    \includegraphics[width=\fwidth]{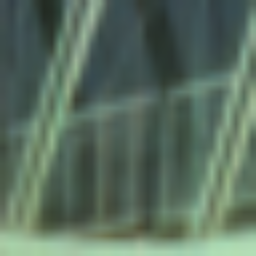}&
     \includegraphics[width=\fwidth]{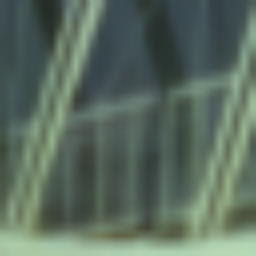}  
     &\includegraphics[width=\fwidth]{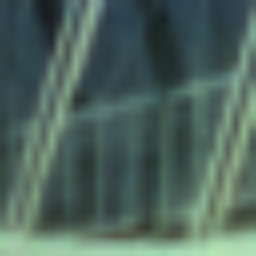}&
     \includegraphics[width=\fwidth]{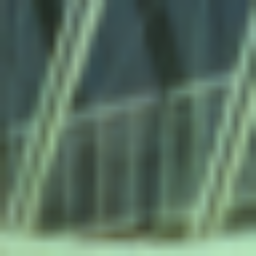} & 
     \includegraphics[width=\fwidth]{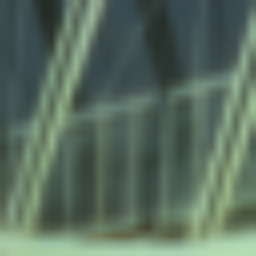}&
      \includegraphics[width=\fwidth]{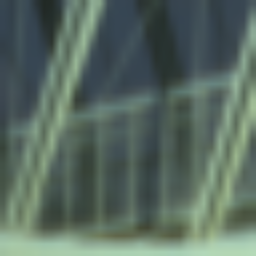} 
     \\ 
      \scriptsize{Noisy}
     & \scriptsize{Noisy-zoom}
      & \scriptsize{QRNN3D}
    & \scriptsize{T3SC}
      &  \scriptsize{GRUNet}
      &\scriptsize{SST}
     & \scriptsize{SERT}
     & \scriptsize{HSDT}
     &  \scriptsize{\textbf{HSDM-B}}
     & \scriptsize{GT}
\end{tabular}
	\caption{Visual comparison of denoising on the ICVL dataset under mixture noise. The HSI bands 29, 19, and 9 combine to create the visual image.}
	\label{fig:icvl_com}
\end{figure*}

\subsection{Experimental Results}

To verify the efficacy of various comparative methods, we adopt a tailored approach. For model-based techniques, we carefully select test configurations that correspond to each model's specific framework. In contrast, deep learning methodologies are evaluated against their publicly accessible pre-trained models, ensuring a fair comparison. The training procedure for all models utilizes the ICVL dataset to maintain consistency across evaluations.

Given the variable number of spectral bands across datasets, it is noteworthy that some comparative models are restricted to data comprising exactly 31 spectral bands. This limitation necessitates a selective application of these models based on dataset compatibility.
To address the challenges posed by datasets with a broader spectral range, we implement a sliding-window strategy for the denoising task.

\subsubsection{Gaussian Noise on Synthetic Datasets}

In this section, we investigate the denoising performance of various methods under fixed Gaussian noise with different standard deviations ($\sigma$) ranging from 10 to 70, applied to the original HSI dataset. Our method is meticulously evaluated against baseline approaches on the ICVL dataset, with the comparative outcomes meticulously detailed in Table \ref{icvl}. Under moderate and low levels of noise interference, our method exhibits performance comparable to SOTA methods. In particular, it demonstrates a denoising advantage when subjected to severe noise disturbances, particularly evident at the noise level of $\sigma=70$.

Figure \ref{fig: icvl} presents the denoising results derived from a variety of methods. A careful examination of these results reveals that methodologies such as LRTDTV, LLRGTV, and MAC-Net, despite their merits, display noticeable pseudo-artifacts in the presence of Gaussian noise. In stark contrast, our proposed HSDM method exemplifies remarkable performance in eliminating these Gaussian noise artifacts.

What sets HSDM apart is its ability to preserve the integrity of intricate details within the HSI dataset, a process deemed crucial for denoising. This not only enhances fidelity but also propels the accuracy to a notch higher compared to other available denoising methods. Manifestly, HSDM is adept at navigating the challenges posed by strong Gaussian noise and preserving intricate dataset details.

\subsubsection{Complex Noise on Synthetic Datasets}

To evaluate the denoising effectiveness in diverse noise environments, we undertake thorough comparative analyses of our approach across multiple datasets, encompassing synthetic datasets ICVL and CAVE and remote sensing datasets like Urban and WDC. Table \ref{icvl_com} delineates the performance metrics, including quantization and noise reduction, specifically for the ICVL dataset under varied noise scenarios. In experiments involving mixture noise variations within the ICVL dataset, HSDM consistently outshines HSDT, showcasing an improvement in PSNR of 0.54 dB.

Furthermore, the results of visual comparison under mixed noise settings on the ICVL dataset are presented in Figure \ref{fig:icvl_com}. It is noteworthy to observe a spectral shift phenomenon in denoising outcomes obtained with GRNet, wherein certain pixels transition from yellow to blue hues. Similarly, T3SC exhibits noticeable color distortion issues. In stark contrast, our proposed MambaHSID model maintains superior spectral consistency and visual fidelity, closely aligning with the GroundTruth across both spectral and spatial domains.

\begin{figure*}[t]
\centering
\includegraphics[width=7in]{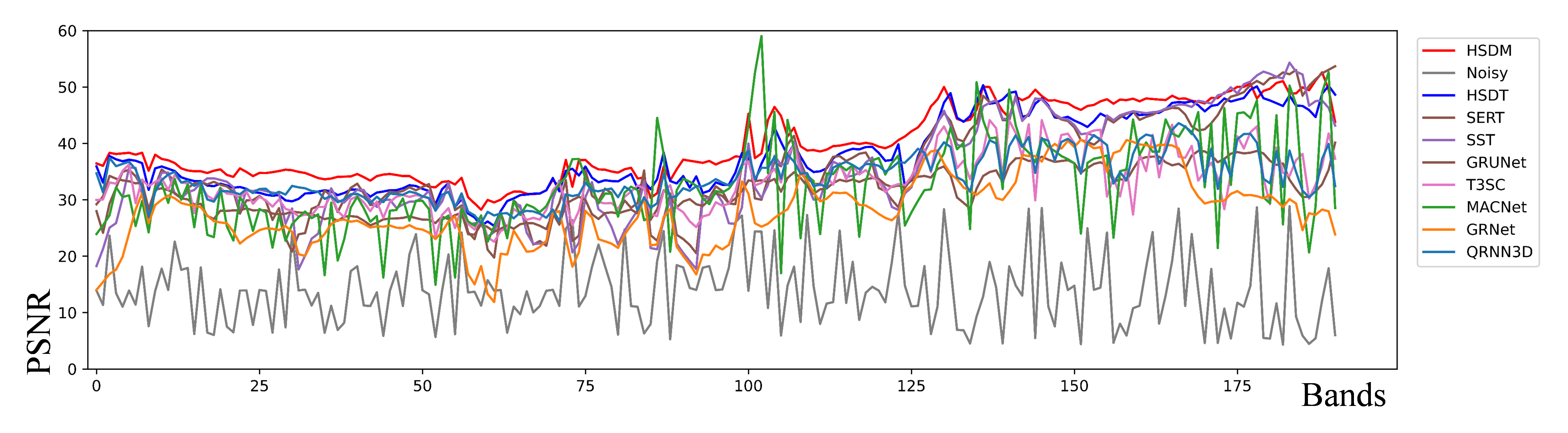}
\caption{Comparative Band-by-Band Analysis of HSDM and Deep-Learning-Based Approaches on the WDC Dataset under Complex Noise Conditions.}\label{wdc_bands}
\end{figure*} 

\begin{figure*}[!t]
    \centering
    \begin{minipage}{0.2\linewidth}
    \centering
\includegraphics[width=0.99\linewidth]{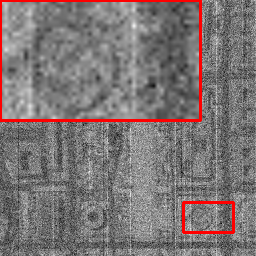}\\
\scriptsize{Noisy}
    \end{minipage}
    \begin{minipage}
    {0.79\linewidth}
    \centering
    % \vspace{-0.2cm}
    \begin{tabular}{ccccc}
     \includegraphics[width=0.19\linewidth]{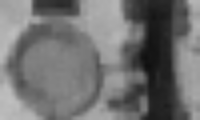} &  
     \includegraphics[width=0.19\linewidth]{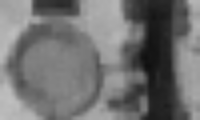}&
     \includegraphics[width=0.19\linewidth]{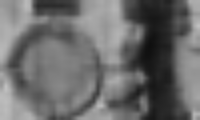}&
     \includegraphics[width=0.19\linewidth]{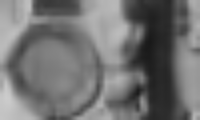}&
     \includegraphics[width=0.19\linewidth]{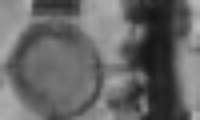} 
     \\
      \scriptsize{LRTDTV}
     & \scriptsize{LLRGTV}
     & \scriptsize{MAC-Net}
     & \scriptsize{T3SC}
     & \scriptsize{QRNN3D}
     \\
     \includegraphics[width=0.19\linewidth]{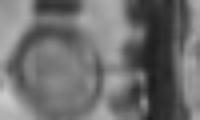} 
     &\includegraphics[width=0.19\linewidth]{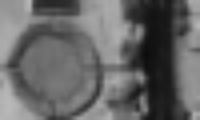}
     &\includegraphics[width=0.19\linewidth]{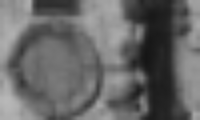}
     &\includegraphics[width=0.19\linewidth]{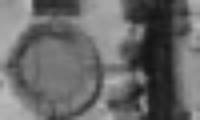}
     &\includegraphics[width=0.19\linewidth]{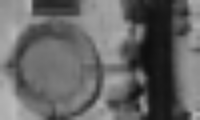} 
    \\  
     \scriptsize{GRUNet}
     &\scriptsize{SST}
     & \scriptsize{SERT}
     & \scriptsize{HSDT}
     &\scriptsize{\textbf{HSDM-B}}
    \end{tabular}
    \end{minipage}
  \caption{Visual comparison of denoising on the WDC dataset band 10 under mixture noise.}
  \label{fig:wdc_10}
\end{figure*}

\begin{figure*}[!t]
    \centering
    \begin{minipage}{0.2\linewidth}
    \centering
\includegraphics[width=0.99\linewidth]{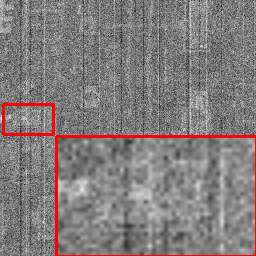}\\
\scriptsize{Noisy}
    \end{minipage}
    \begin{minipage}
    {0.79\linewidth}
    \centering
    % \vspace{-0.2cm}
    \begin{tabular}{ccccc}
     \includegraphics[width=0.19\linewidth]{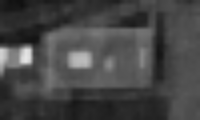} &  
     \includegraphics[width=0.19\linewidth]{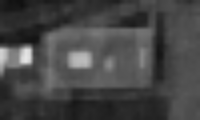}&
     \includegraphics[width=0.19\linewidth]{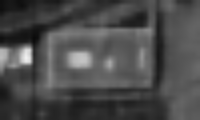}&
     \includegraphics[width=0.19\linewidth]{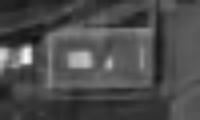}&
     \includegraphics[width=0.19\linewidth]{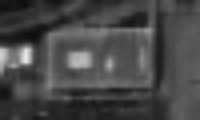} 
     \\
      \scriptsize{LRTDTV}
     & \scriptsize{LLRGTV}
     & \scriptsize{MAC-Net}
     & \scriptsize{T3SC}
     & \scriptsize{QRNN3D}
     \\
     \includegraphics[width=0.19\linewidth]{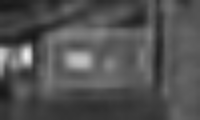} 
     &\includegraphics[width=0.19\linewidth]{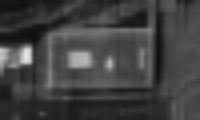}
     &\includegraphics[width=0.19\linewidth]{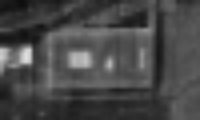}
     &\includegraphics[width=0.19\linewidth]{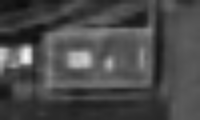}
     &\includegraphics[width=0.19\linewidth]{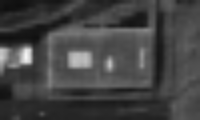} 
    \\  
     \scriptsize{GRUNet}
     &\scriptsize{SST}
     & \scriptsize{SERT}
     & \scriptsize{HSDT}
     &\scriptsize{\textbf{HSDM-B}}
    \end{tabular}
    \end{minipage}
  \caption{Visual comparison of denoising on the WDC dataset band 76 under mixture noise.}
  \label{fig:wdc_76}
\end{figure*}

% \def\fwidth{0.19\linewidth}
% \def\arraystretch{0.5}
% \renewcommand{\tabcolsep}{0.5 pt}
% \begin{figure*}[t]
% \centering
% \begin{tabular}{ccccc}
%      \includegraphics[width=\fwidth]{result/urban/noisy.png}&
%      \includegraphics[width=\fwidth]{result/urban/LRTDTV.png} &  
%      \includegraphics[width=\fwidth]{result/urban/LLRGTV.png}&
%      \includegraphics[width=\fwidth]{result/urban/macnet.png}&
%      \includegraphics[width=\fwidth]{result/urban/t3sc.png}
%      \\
%             \scriptsize{Noisy}
%      & \scriptsize{LRTDTV}
%      & \scriptsize{LLRGTV}
%      & \scriptsize{MAC-Net}
%      & \scriptsize{T3SC}
%      \\
%      \includegraphics[width=\fwidth]{result/urban/qrnn3d.png} &  
%      \includegraphics[width=\fwidth]{result/urban/sst.png}&
%      \includegraphics[width=\fwidth]{result/urban/sert_base.png}&
%      \includegraphics[width=\fwidth]{result/urban/hsdt.png}&
%      \includegraphics[width=\fwidth]{result/urban/hsid_mamba.png} 
%     \\
%       \scriptsize{QRNN3D}
%      & \scriptsize{SST}
%      & \scriptsize{SERT}
%      & \scriptsize{HSDT}
%      &\scriptsize{\textbf{HSDM}}
% \end{tabular}
%     \caption{Visual comparison of denoising on the Urban dataset. The HSI bands 102,138 and 202 combine to create the visual pseudocolor image. }
% 	\label{fig:urban}
% \end{figure*}

\begin{figure*}[!t]
    \centering
    \begin{minipage}{0.2\linewidth}
    \centering
\includegraphics[width=0.99\linewidth]{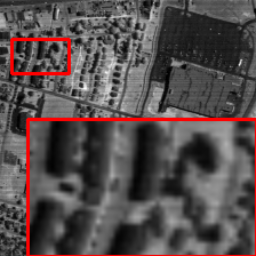}\\
\scriptsize{Noisy}
    \end{minipage}
    \begin{minipage}
    {0.79\linewidth}
    \centering
    % \vspace{-0.2cm}
    \begin{tabular}{ccccc}
     \includegraphics[width=0.19\linewidth]{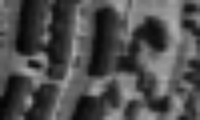} &  
     \includegraphics[width=0.19\linewidth]{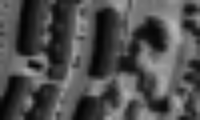}&
     \includegraphics[width=0.19\linewidth]{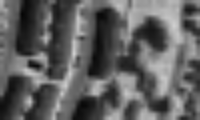}&
     \includegraphics[width=0.19\linewidth]{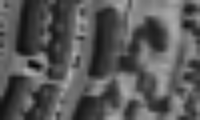}&
     \includegraphics[width=0.19\linewidth]{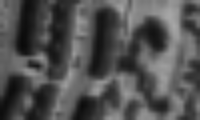} 
     \\
      \scriptsize{LRTDTV}
     & \scriptsize{LLRGTV}
     & \scriptsize{MAC-Net}
     & \scriptsize{T3SC}
     & \scriptsize{QRNN3D}
     \\
     \includegraphics[width=0.19\linewidth]{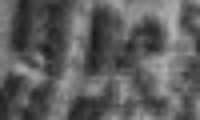} 
     &\includegraphics[width=0.19\linewidth]{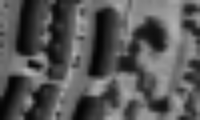}
     &\includegraphics[width=0.19\linewidth]{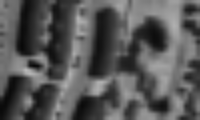}
     &\includegraphics[width=0.19\linewidth]{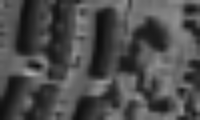}
     &\includegraphics[width=0.19\linewidth]{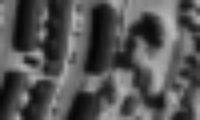} 
    \\  
     \scriptsize{GRUNet}
     &\scriptsize{SST}
     & \scriptsize{SERT}
     & \scriptsize{HSDT}
     &\scriptsize{\textbf{HSDM-B}}
    \end{tabular}
    \end{minipage}
  \caption{Visual comparison of denoising on the Urban dataset band 102 under mixture noise.}
  \label{fig:urban_102}
\end{figure*}

\begin{figure*}[!t]
    \centering
    \begin{minipage}{0.2\linewidth}
    \centering
\includegraphics[width=0.99\linewidth]{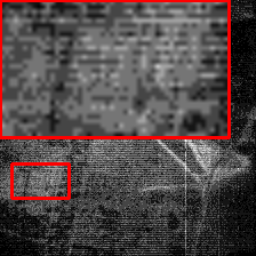}\\
\scriptsize{Noisy}
    \end{minipage}
    \begin{minipage}
    {0.79\linewidth}
    \centering
    % \vspace{-0.2cm}
    \begin{tabular}{ccccc}
     \includegraphics[width=0.19\linewidth]{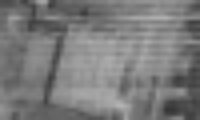} &  
     \includegraphics[width=0.19\linewidth]{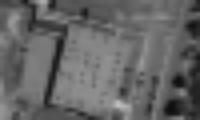}&
     \includegraphics[width=0.19\linewidth]{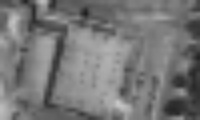}&
     \includegraphics[width=0.19\linewidth]{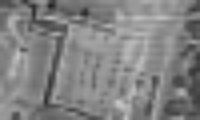}&
     \includegraphics[width=0.19\linewidth]{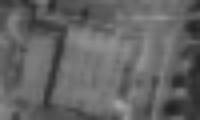} 
     \\
      \scriptsize{LRTDTV}
     & \scriptsize{LLRGTV}
     & \scriptsize{MAC-Net}
     & \scriptsize{T3SC}
     & \scriptsize{QRNN3D}
     \\
     \includegraphics[width=0.19\linewidth]{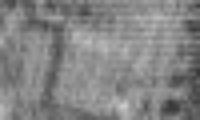} 
     &\includegraphics[width=0.19\linewidth]{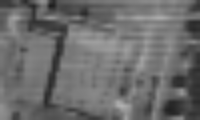}
     &\includegraphics[width=0.19\linewidth]{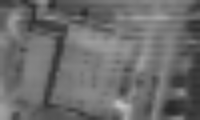}
     &\includegraphics[width=0.19\linewidth]{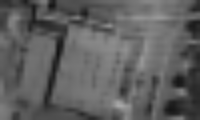}
     &\includegraphics[width=0.19\linewidth]{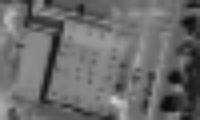} 
    \\  
     \scriptsize{GRUNet}
     &\scriptsize{SST}
     & \scriptsize{SERT}
     & \scriptsize{HSDT}
     &\scriptsize{\textbf{HSDM-B}}
    \end{tabular}
    \end{minipage}
  \caption{Visual comparison of denoising on the Urban dataset band 138 under mixture noise.}
  \label{fig:urban_138}
\end{figure*}

\begin{figure*}[!t]
    \centering
    \begin{minipage}{0.2\linewidth}
    \centering
\includegraphics[width=0.99\linewidth]{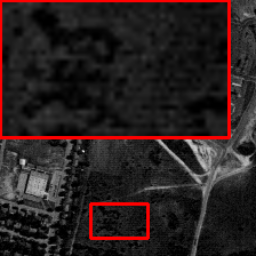}\\
\scriptsize{Noisy}
    \end{minipage}
    \begin{minipage}
    {0.79\linewidth}
    \centering
    % \vspace{-0.2cm}
    \begin{tabular}{ccccc}
     \includegraphics[width=0.19\linewidth]{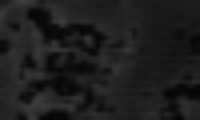} &  
     \includegraphics[width=0.19\linewidth]{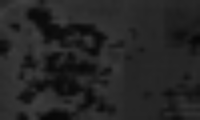}&
     \includegraphics[width=0.19\linewidth]{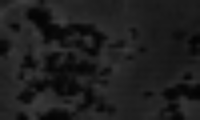}&
     \includegraphics[width=0.19\linewidth]{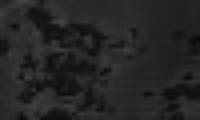}&
     \includegraphics[width=0.19\linewidth]{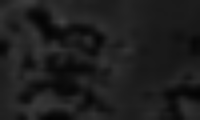} 
     \\
      \scriptsize{LRTDTV}
     & \scriptsize{LLRGTV}
     & \scriptsize{MAC-Net}
     & \scriptsize{T3SC}
     & \scriptsize{QRNN3D}
     \\
     \includegraphics[width=0.19\linewidth]{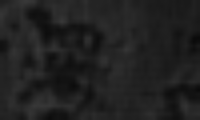} 
     &\includegraphics[width=0.19\linewidth]{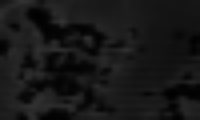}
     &\includegraphics[width=0.19\linewidth]{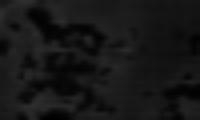}
     &\includegraphics[width=0.19\linewidth]{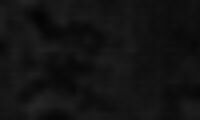}
     &\includegraphics[width=0.19\linewidth]{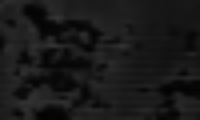} 
    \\  
     \scriptsize{GRUNet}
     &\scriptsize{SST}
     & \scriptsize{SERT}
     & \scriptsize{HSDT}
     &\scriptsize{\textbf{HSDM-B}}
    \end{tabular}
    \end{minipage}
  \caption{Visual comparison of denoising on the Urban dataset band 202 under mixture noise.}
  \label{fig:urban_202}
\end{figure*}

As shown in Figure~\ref{wdc_bands}, the comprehensive comparative analysis delineated through WDCMall under mixture noise shows that HSDM presents superior performance in a broad spectrum of bands. Figure~\ref{fig:wdc_10} and Figure~\ref{fig:wdc_76} elucidate the comparative efficacy of different bands, where ours consistently emerges as the most effective, indicating superior handling of variability and noise in the data.

\subsubsection{Noise on Real World}

To validate the model's robustness under real-world noise conditions, we conducted hyperspectral image denoising experiments on representative datasets Urban. Figures Figure~\ref{fig:urban_102}, Figure~\ref{fig:urban_138}, and Figure~\ref{fig:urban_202} illustrate the visual denoising performance across different bands of Urban datasets. Visual denoising performance is strikingly portrayed. In particular, our method demonstrates superior effectiveness, particularly in detail preservation and intensity transition regions. It meaningfully maintains the integrity of the details, subtly managing the noise and offering smoother transitions between diverse intensity levels. This superior demarcation of details during denoising and the exceptional performance at brightness transition zones underline the precision and robustness of our methodology.

Moreover, we extended our denoising experiments to include the CAVE dataset and the quantitative results are detailed in Table \ref{module_effcient}. HSDM showed remarkable superiority in real noise removal, as evidenced by its higher PSNR and SSIM values compared to other methods. Although LRTDTV and LLRGTV demonstrated a certain degree of alignment with the SAM metric changes on real datasets due to their priori global low-rank assumptions, their performance was relatively poorer on additional datasets. In contrast, HSDM consistently exhibited superior performance across a wider range of datasets, underscoring its versatility and adaptability to various HSI denoising scenarios.

\subsection{Ablation experiment}

\subsubsection{Effectiveness of scanning mechanism}

To better explore the impact of scanning mechanisms combined with SSM on HSI denoising, we conducted experiments using both popular Sweep scanning and continuous scanning methods on different datasets. We performed both unidirectional and bidirectional experiments for each scanning method, and the results are summarized in Table \ref{module_effcient}, Alt representing whether to use the alternate mechanism. The findings provide evidence supporting the effectiveness of our bidirectional Continuous scanning approach.

\renewcommand{\tabcolsep}{3pt}
\renewcommand\arraystretch{1.3}
\begin{table}[t]
    \caption{Ablation study on the scanning mechanism on three datasets under mixture noise.}
    \centering
    \begin{tabular}{c|c|c|cc|cc|cc}
    \toprule
         \multirow{2}{*}{Baseline}&\multirow{2}{*}{Alt}&  \multirow{2}{*}{Params}& \multicolumn{2}{c|}{ICVL}&\multicolumn{2}{c|}{CAVE}&\multicolumn{2}{c}{WDC}\\
         % &GFLOPS 
         &&&PSNR&SSIM&PSNR&SSIM&PSNR&SSIM\\
         \midrule
          Sweep &-&0.93 &37.15&0.9883&34.72&0.9802&36.46&0.9868\\
           Continuous &-&0.93 &37.58&0.9870&34.89&0.9657&36.72&0.9846\\
          \midrule 
         Sweep &\checkmark&0.68 &38.73&0.9911&35.19&0.9783&36.88&0.9872\\
         Bid-Sweep &\checkmark& 0.68 &39.61&0.9936&35.42&0.9782&37.24&0.9881\\
         Continuous &\checkmark&0.68&39.17&0.9884&36.38&0.9832&37.19& 0.9871\\
         \rowcolor{gray9}
         Bid-Cont &\checkmark& 0.68 &41.30&0.9950 &35.99&0.9867&37.57&0.9890\\
         \bottomrule
    \end{tabular}

    \label{module_effcient}
\end{table}

Despite the varying number of individual scan iterations required by different scanning mechanisms in the comparative experiments, we ensure uniformity in the total number of scans by selecting a specific scanning strategy at each layer, thereby maintaining a consistent model size across all variations.

The results demonstrate that our bidirectional continuous scanning approach outperforms other scanning methods, indicating its effectiveness in capturing complex spatial-spectral relationships within the hyperspectral data. This finding underscores the importance of considering bidirectional scanning mechanisms when implementing state space models for HSI denoising.

\subsubsection{Effectiveness of modifications}
                                              
To enhance the effectiveness of the original Mamba module for HSI denoising, we have introduced several enhancements. Firstly, we have incorporated Conv1D operations to optimize the processing of linear sequences, thereby enhancing the model's capacity to capture both spatial and spectral dependencies. Furthermore, we have integrated channel attention mechanisms specifically designed for HSI, aimed at further enriching the capabilities of the Mamba framework. The results of these enhancements are summarized in Table \ref{window_effcient}.

\renewcommand{\tabcolsep}{3pt}
\renewcommand\arraystretch{1.3}
\begin{table}[t]
    \caption{Ablation study of different attention window mechanisms on ICVL dataset under mixture noise.}
    \centering
    \begin{tabular}{c|ccc}
    \toprule
         Method&PSNR(dB) &SSIM &SAM\\
    \midrule
        % \rowcolor{gray9}
         Baseline   &41.30&0.9950&0.0459\\
         Baseline + w/o Conv1D &41.09&0.9945&0.0293\\
         Baseline + w/o Channel Attention &39.87&0.9934&0.0351\\
    \bottomrule
    \end{tabular}

    \label{window_effcient}
\end{table}

\subsubsection{Comparison with other vision mamba methods} In the chapter on dissolving experiment in our paper, we allocate a section to evaluate the performance of our proposition, HSDM, in comparison to the similar data-based VideoMamba and another similar denoising model, MambaIR. Unlike these two methodologies, HSDM employs a customized scanning architecture specifically designed for HSI and a pixel-smoothing scanning technique. This unique approach enables HSDM to achieve superior HSI denoising results at a more rapid pace shown in Table \ref{mamba_effcient}. Therefore, the proposed HSDM offers a compelling and efficient alternative in the HSI denoising domain, effectively dealing with challenges posed by other models while providing superior performance and speed in retrieval.

\renewcommand{\tabcolsep}{3pt}
\renewcommand\arraystretch{1.3}
\begin{table}[h]
    \caption{Ablation study of different spectral enhancement strategies on ICVL dataset under mixture noise.}
    \centering
    \begin{tabular}{c|c|c|ccc}
    \toprule
         hidden dim& Params(M) & Time(sec) &PSNR(dB) &SSIM &SAM\\
    \midrule
         MambaIR~\cite{mambair}&4.35&2.24&39.80&0.9900&0.0401\\
         VideoMamba~\cite{videomamba}&1.28&1.43&39.62&0.9843&0.0364\\
         \rowcolor{gray9}
         HSDM&0.68&0.77&41.30&0.9950&0.0459\\
    \bottomrule
    \end{tabular}

    \label{architecture_effcient}
\end{table}

\subsubsection{Comparison with other architecture methods} To further elucidate the efficacy of our proposed HSDM architecture, we present a comprehensive comparison including train time(sec), inference time(sec), and GPU usage with GrUNet and Transformer-based methods on a 3090 GPU, the results are shown in Figure~\ref{architecture_effcient}. The higher PSNR values of HSDM highlight its superior spatial-spectral utilization for HSI-denoising tasks. Enhanced computational efficiency and optimized utilization of GPU resources further position HSDM as an effective choice for comprehensive hyperspectral imaging endeavors. Thus, our HSDM validates the potency of the Mamba architecture within the HSI denoising domain.

\renewcommand{\tabcolsep}{3pt}
\renewcommand\arraystretch{1.3}
\begin{table}[h]
    \caption{Ablation study of different architecture methods on CAVE dataset under mixture noise. SST and SERT are transformer-based methods.}
    \centering
    \begin{tabular}{c|c|c|c|c|c}
    \toprule
         Methods& Params(M) &Train& Inference& Usage&PSNR \\
    \midrule
         GRUNet~\cite{GRUNet}&41.44&0.27&0.63&9824&28.44\\
         SST~\cite{sst}&4.10 &0.46&0.34&11348&34.89\\
         SERT~\cite{sert}&1.91&0.28&1.29&5104&35.86\\
         \rowcolor{gray9}
         HSDM&0.68&0.36&0.77&4644&41.30\\
    \bottomrule
    \end{tabular}

    \label{mamba_effcient}
\end{table}

\subsection{Parameter analysis}

HSDM consistently maintains its GPU memory consumption and FPS rate. In contrast, the primary baseline Bidirectional HSDM experiences a decrease in FPS due to increased parameters. Notably, the impact of varying numbers of hidden channels on the model's feature-capturing ability is significant. We analyze different numbers of hidden channels, presented in Table \ref{SE_effcient}. This decision strikes a better balance between the model's efficiency and performance considerations. By increasing the number of hidden channels to 48, the model gains an enhanced ability to capture and represent complex features within the data. Thus, \colorbox{blue!25}{ } represents the HSDM-B version that we ultimately selected, while \colorbox{red!25}{  } represent the HSDM-L version. This decision contributes to improved model performance without excessively increasing computational complexity, thus optimizing the trade-off between efficiency and effectiveness.

\renewcommand{\tabcolsep}{3pt}
\renewcommand\arraystretch{1.3}
\begin{table}[t]
    \caption{Ablation study of different spectral enhancement strategies on ICVL dataset under mixture noise.}
    \centering
    \begin{tabular}{c|ccc|ccc}
    \toprule
         Hidden dim& Params(M) &Runtime & Usage &PSNR(dB) &SSIM &SAM\\
    \midrule
         24&0.24&0.58&2896&38.85&0.9901&0.0418\\
         32&0.44&0.67&3766&39.71&0.9936&0.0350\\
         \rowcolor{blue!25}
         48&0.68&0.79&4644&41.30&0.9950&0.0459\\
         56&0.86&0.83&5282&41.50&0.9937&0.0374\\
         \rowcolor{red!25}
         64&1.08&0.90&5876&41.62&0.9954&0.0431\\
         
    \bottomrule
    \end{tabular}

    \label{SE_effcient}
\end{table}

\section{Conclusion}

In this paper, we present the Hyperspectral Image Denoising Mamba-based network (HSDM), an innovative HSI denoising approach grounded in the selective state space model framework. HSDM employs a novel selective state-space model and a unique Hyperspectral Continuous Scanning Block (HCSB) for advanced denoising. HSDM effectively addresses the limitations of convolutional and attention mechanisms, leveraging an advanced 2D Selective Scan Module for spatial-spectral dependencies and maintaining linear complexity. The alternating continuous scanning mechanism improves efficiency and minimizes distortions. With the integration of residual learning and spectral attention, HSDM achieves exceptional noise reduction while preserving essential image features. Extensive experimental validations are conducted on diverse datasets, HSDM surpasses existing methods, highlighting its effectiveness and efficiency in tackling the challenges of HSI denoising.

\bibliographystyle{IEEEtran}
\bibliography{IEEEabrv,reference}

\begin{thebibliography}{10}
\providecommand{\url}[1]{#1}
\csname url@samestyle\endcsname
\providecommand{\newblock}{\relax}
\providecommand{\bibinfo}[2]{#2}
\providecommand{\BIBentrySTDinterwordspacing}{\spaceskip=0pt\relax}
\providecommand{\BIBentryALTinterwordstretchfactor}{4}
\providecommand{\BIBentryALTinterwordspacing}{\spaceskip=\fontdimen2\font plus
\BIBentryALTinterwordstretchfactor\fontdimen3\font minus \fontdimen4\font\relax}
\providecommand{\BIBforeignlanguage}[2]{{%
\expandafter\ifx\csname l@#1\endcsname\relax
\typeout{** WARNING: IEEEtran.bst: No hyphenation pattern has been}%
\typeout{** loaded for the language `#1'. Using the pattern for}%
\typeout{** the default language instead.}%
\else
\language=\csname l@#1\endcsname
\fi
#2}}
\providecommand{\BIBdecl}{\relax}
\BIBdecl

\bibitem{aburaed2023review}
N.~Aburaed, M.~Q. Alkhatib, S.~Marshall, J.~Zabalza, and H.~Al~Ahmad, ``A review of spatial enhancement of hyperspectral remote sensing imaging techniques,'' \emph{IEEE Journal of Selected Topics in Applied Earth Observations and Remote Sensing}, vol.~16, pp. 2275--2300, 2023.

\bibitem{he2023object}
X.~He, C.~Tang, X.~Liu, W.~Zhang, K.~Sun, and J.~Xu, ``Object detection in hyperspectral image via unified spectral-spatial feature aggregation,'' \emph{IEEE Transactions on Geoscience and Remote Sensing}, 2023.

\bibitem{jang2024m}
J.~Jang, S.~Oh, Y.~Kim, D.~Seo, Y.~Choi, and H.~J. Yang, ``Sodai: Multi-modal maritime object detection dataset with rgb and hyperspectral image sensors,'' \emph{Advances in Neural Information Processing Systems}, vol.~36, 2024.

\bibitem{yang2023iter}
J.~Yang, B.~Du, D.~Wang, and L.~Zhang, ``Iter: Image-to-pixel representation for weakly supervised hsi classification,'' \emph{IEEE Transactions on Image Processing}, 2023.

\bibitem{ding2023multi}
Y.~Ding, Z.~Zhang, X.~Zhao, D.~Hong, W.~Cai, N.~Yang, and B.~Wang, ``Multi-scale receptive fields: Graph attention neural network for hyperspectral image classification,'' \emph{Expert Systems with Applications}, vol. 223, p. 119858, 2023.

\bibitem{sun2018hyperspectral}
L.~Sun and B.~Jeon, ``Hyperspectral mixed denoising via subspace low rank learning and bm4d filtering,'' in \emph{IGARSS 2018-2018 IEEE International Geoscience and Remote Sensing Symposium}.\hskip 1em plus 0.5em minus 0.4em\relax IEEE, 2018, pp. 8034--8037.

\bibitem{LLRGTV}
W.~He, H.~Zhang, H.~Shen, and L.~Zhang, ``Hyperspectral image denoising using local low-rank matrix recovery and global spatial--spectral total variation,'' \emph{IEEE Journal of Selected Topics in Applied Earth Observations and Remote Sensing}, vol.~11, no.~3, pp. 713--729, 2018.

\bibitem{NG-Meet}
W.~He, Q.~Yao, C.~Li, N.~Yokoya, and Q.~Zhao, ``Non-local meets global: An integrated paradigm for hyperspectral denoising,'' 2019, pp. 6868--6877.

\bibitem{guan2022dnrcnn}
J.~Guan, R.~Lai, H.~Li, Y.~Yang, and L.~Gu, ``Dnrcnn: Deep recurrent convolutional neural network for hsi destriping,'' \emph{IEEE Transactions on Neural Networks and Learning Systems}, 2022.

\bibitem{qrnn3d}
K.~Wei, Y.~Fu, and H.~Huang, ``3-d quasi-recurrent neural network for hyperspectral image denoising,'' vol.~32, no.~1, pp. 363--375, 2020.

\bibitem{khan2022transformers}
S.~Khan, M.~Naseer, M.~Hayat, S.~W. Zamir, F.~S. Khan, and M.~Shah, ``Transformers in vision: A survey,'' \emph{ACM computing surveys (CSUR)}, vol.~54, no. 10s, pp. 1--41, 2022.

\bibitem{sst}
M.~Li, Y.~Fu, and Y.~Zhang, ``Spatial-spectral transformer for hyperspectral image denoising,'' vol.~37, no.~1, 2023, pp. 1368--1376.

\bibitem{sert}
M.~Li, J.~Liu, Y.~Fu, Y.~Zhang, and D.~Dou, ``Spectral enhanced rectangle transformer for hyperspectral image denoising,'' 2023, pp. 5805--5814.

\bibitem{s4}
A.~Gu, K.~Goel, and C.~R{\'e}, ``Efficiently modeling long sequences with structured state spaces,'' \emph{arXiv preprint arXiv:2111.00396}, 2021.

\bibitem{S5}
J.~Smith, A.~Warrington, and S.~Linderman, ``\BIBforeignlanguage{en-US}{Simplified state space layers for sequence modeling},'' Aug 2022.

\bibitem{H3}
D.~Y. Fu, T.~Dao, K.~K. Saab, A.~W. Thomas, A.~Rudra, and C.~R{\'e}, ``Hungry hungry hippos: Towards language modeling with state space models,'' \emph{arXiv preprint arXiv:2212.14052}, 2022.

\bibitem{GSS}
H.~Mehta, A.~Gupta, A.~Cutkosky, and B.~Neyshabur, ``\BIBforeignlanguage{en-US}{Long range language modeling via gated state spaces},'' Jun 2022.

\bibitem{Mamba}
A.~Gu and T.~Dao, ``\BIBforeignlanguage{en-US}{Mamba: Linear-time sequence modeling with selective state spaces},'' Dec 2023.

\bibitem{videomamba}
K.~Li, X.~Li, Y.~Wang, Y.~He, Y.~Wang, L.~Wang, and Y.~Qiao, ``Videomamba: State space model for efficient video understanding,'' \emph{arXiv preprint arXiv:2403.06977}, 2024.

\bibitem{mambair}
H.~Guo, J.~Li, T.~Dai, Z.~Ouyang, X.~Ren, and S.-T. Xia, ``Mambair: A simple baseline for image restoration with state-space model,'' \emph{arXiv preprint arXiv:2402.15648}, 2024.

\bibitem{zigmamba}
V.~T. Hu, S.~A. Baumann, M.~Gui, O.~Grebenkova, P.~Ma, J.~Fischer, and B.~Ommer, ``Zigma: Zigzag mamba diffusion model,'' \emph{arXiv preprint arXiv:2403.13802}, 2024.

\bibitem{visionmamba}
L.~Zhu, B.~Liao, Q.~Zhang, X.~Wang, W.~Liu, and X.~Wang, ``Vision mamba: Efficient visual representation learning with bidirectional state space model,'' \emph{arXiv preprint arXiv:2401.09417}, 2024.

\bibitem{zhang2019hybrid}
Q.~Zhang, Q.~Yuan, J.~Li, X.~Liu, H.~Shen, and L.~Zhang, ``Hybrid noise removal in hyperspectral imagery with a spatial--spectral gradient network,'' \emph{IEEE Transactions on Geoscience and Remote Sensing}, vol.~57, no.~10, pp. 7317--7329, 2019.

\bibitem{yuan2018hyperspectral}
Q.~Yuan, Q.~Zhang, J.~Li, H.~Shen, and L.~Zhang, ``Hyperspectral image denoising employing a spatial--spectral deep residual convolutional neural network,'' \emph{IEEE Transactions on Geoscience and Remote Sensing}, vol.~57, no.~2, pp. 1205--1218, 2018.

\bibitem{TRQ3DNet}
L.~Pang, W.~Gu, and X.~Cao, ``Trq3dnet: A 3d quasi-recurrent and transformer based network for hyperspectral image denoising,'' \emph{Remote Sensing}, vol.~14, no.~18, p. 4598, 2022.

\bibitem{sstd}
M.~Wang, W.~He, and H.~Zhang, ``A spatial–spectral transformer network with total variation loss for hyperspectral image denoising,'' \emph{IEEE Geoscience and Remote Sensing Letters}, vol.~20, pp. 1--5, 2023.

\bibitem{hsimamba}
J.~X. Yang, J.~Zhou, J.~Wang, H.~Tian, and A.~W.~C. Liew, ``Hsimamba: Hyperpsectral imaging efficient feature learning with bidirectional state space for classification,'' \emph{arXiv preprint arXiv:2404.00272}, 2024.

\bibitem{medmamba}
Y.~Yue and Z.~Li, ``Medmamba: Vision mamba for medical image classification,'' \emph{arXiv preprint arXiv:2403.03849}, 2024.

\bibitem{LRTDTV}
Y.~Wang, J.~Peng, Q.~Zhao, Y.~Leung, X.-L. Zhao, and D.~Meng, ``Hyperspectral image restoration via total variation regularized low-rank tensor decomposition,'' \emph{IEEE Journal of Selected Topics in Applied Earth Observations and Remote Sensing}, vol.~11, no.~4, pp. 1227--1243, 2017.

\bibitem{GRN}
X.~Cao, X.~Fu, C.~Xu, and D.~Meng, ``Deep spatial-spectral global reasoning network for hyperspectral image denoising,'' vol.~60, pp. 1--14, 2021.

\bibitem{xiong2021mac}
F.~Xiong, J.~Zhou, Q.~Zhao, J.~Lu, and Y.~Qian, ``Mac-net: Model-aided nonlocal neural network for hyperspectral image denoising,'' \emph{IEEE Transactions on Geoscience and Remote Sensing}, vol.~60, pp. 1--14, 2021.

\bibitem{bodrito2021trainable}
T.~Bodrito, A.~Zouaoui, J.~Chanussot, and J.~Mairal, ``A trainable spectral-spatial sparse coding model for hyperspectral image restoration,'' \emph{Advances in Neural Information Processing Systems}, vol.~34, pp. 5430--5442, 2021.

\bibitem{GRUNet}
Z.~Lai, K.~Wei, and Y.~Fu, ``Deep plug-and-play prior for hyperspectral image restoration,'' \emph{Neurocomputing}, vol. 481, pp. 281--293, 2022.

\bibitem{man}
Z.~Lai and Y.~Fu, ``Mixed attention network for hyperspectral image denoising,'' \emph{arXiv preprint arXiv:2301.11525}, 2023.

\bibitem{li2023spatial}
M.~Li, Y.~Fu, and Y.~Zhang, ``Spatial-spectral transformer for hyperspectral image denoising,'' in \emph{Proceedings of the AAAI Conference on Artificial Intelligence}, vol.~37, no.~1, 2023, pp. 1368--1376.

\bibitem{li2023spectral}
M.~Li, J.~Liu, Y.~Fu, Y.~Zhang, and D.~Dou, ``Spectral enhanced rectangle transformer for hyperspectral image denoising,'' in \emph{Proceedings of the IEEE/CVF Conference on Computer Vision and Pattern Recognition}, 2023, pp. 5805--5814.

\bibitem{hsdt}
Z.~Lai, C.~Yan, and Y.~Fu, ``Hybrid spectral denoising transformer with guided attention,'' 2023, pp. 13\,065--13\,075.

\bibitem{wei20203}
K.~Wei, Y.~Fu, and H.~Huang, ``3-d quasi-recurrent neural network for hyperspectral image denoising,'' \emph{IEEE transactions on neural networks and learning systems}, vol.~32, no.~1, pp. 363--375, 2020.

\bibitem{arad2016sparse}
B.~Arad and O.~Ben-Shahar, ``Sparse recovery of hyperspectral signal from natural rgb images,'' in \emph{Computer Vision--ECCV 2016: 14th European Conference, Amsterdam, The Netherlands, October 11--14, 2016, Proceedings, Part VII 14}.\hskip 1em plus 0.5em minus 0.4em\relax Springer, 2016, pp. 19--34.

\bibitem{CAVE}
J.-I. Park, M.-H. Lee, M.~D. Grossberg, and S.~K. Nayar, ``Multispectral imaging using multiplexed illumination,'' 2007, pp. 1--8.

\bibitem{urban}
V.~Mnih and G.~E. Hinton, ``Learning to detect roads in high-resolution aerial images.''\hskip 1em plus 0.5em minus 0.4em\relax Springer, 2010, pp. 210--223.

\bibitem{kuang2019fashion}
Z.~Kuang, Y.~Gao, G.~Li, P.~Luo, Y.~Chen, L.~Lin, and W.~Zhang, ``Fashion retrieval via graph reasoning networks on a similarity pyramid,'' in \emph{Proceedings of the IEEE/CVF international conference on computer vision}, 2019, pp. 3066--3075.

\bibitem{SSIM}
Z.~Wang, A.~C. Bovik, H.~R. Sheikh, and E.~P. Simoncelli, ``Image quality assessment: from error visibility to structural similarity,'' \emph{IEEE transactions on image processing}, vol.~13, no.~4, pp. 600--612, 2004.

\bibitem{SAM}
R.~H. Yuhas, J.~W. Boardman, and A.~F. Goetz, ``Determination of semi-arid landscape endmembers and seasonal trends using convex geometry spectral unmixing techniques,'' in \emph{JPL, Summaries of the 4th Annual JPL Airborne Geoscience Workshop. Volume 1: AVIRIS Workshop}, 1993.

\bibitem{MACNet}
F.~Xiong, J.~Zhou, Q.~Zhao, J.~Lu, and Y.~Qian, ``Mac-net: Model-aided nonlocal neural network for hyperspectral image denoising,'' vol.~60, pp. 1--14, 2021.

\bibitem{T3SC}
T.~Bodrito, A.~Zouaoui, J.~Chanussot, and J.~Mairal, ``A trainable spectral-spatial sparse coding model for hyperspectral image restoration,'' vol.~34, pp. 5430--5442, 2021.

\bibitem{adam}
D.~P. Kingma and J.~Ba, ``Adam: A method for stochastic optimization,'' \emph{arXiv preprint arXiv:1412.6980}, 2014.

\bibitem{LLRT}
Y.~Chang, L.~Yan, and S.~Zhong, ``Hyper-laplacian regularized unidirectional low-rank tensor recovery for multispectral image denoising,'' 2017, pp. 4260--4268.

\end{thebibliography}

% \end{thebibliography}

% \newpage

 

% \end{IEEEbiography}



\vfill

\end{document}